\documentclass{article} 
\usepackage[table,xcdraw]{xcolor}
\usepackage{iclr2024_conference,times}

\usepackage{amsmath,amsfonts,bm}

\def\eqref#1{equation~\ref{#1}}

\def\1{\bm{1}}

\DeclareMathAlphabet{\mathsfit}{\encodingdefault}{\sfdefault}{m}{sl}
\SetMathAlphabet{\mathsfit}{bold}{\encodingdefault}{\sfdefault}{bx}{n}

\usepackage{hyperref}
\usepackage{url}

\usepackage{bbm}
\usepackage{booktabs}
\usepackage{amsmath}
\usepackage{xcolor}
\usepackage{multirow}
\usepackage{amssymb}
\usepackage{bbding}
\usepackage{pifont}
\usepackage{utfsym}
\usepackage{wrapfig}

\definecolor{GEN}{RGB}{255,165,0}
\definecolor{CTX}{RGB}{169,169,169}
\definecolor{RET}{RGB}{128,0,128}
\newcommand{\ours}{OneGen}
\usepackage{algorithm}
\usepackage{algorithmicx}
\usepackage{algpseudocode}
\floatname{algorithm}{Algorithm} 
\usepackage{color}
\usepackage{caption}
\usepackage{tcolorbox}
\definecolor{uclablue}{rgb}{0.15, 0.45, 0.68}
\hypersetup{
    breaklinks,
    citecolor=uclablue,
    colorlinks=true,
    linkcolor=uclablue
}
\usepackage{calc}
\newlength\myheight
\newlength\mydepth
\settototalheight\myheight{Xygp}
\usepackage{titletoc}

\title{OneGen: Efficient One-Pass Unified \\ Generation and Retrieval for LLMs}

\author{
  Jintian Zhang${^{\spadesuit\heartsuit}\footnotemark[1]}$~, 
  Cheng Peng${^{\heartsuit\clubsuit}}$\thanks{$\quad$ Equal Contribution.}~, 
  Mengshu Sun$^{\heartsuit\clubsuit}$, 
  Xiang Chen$^{\spadesuit\heartsuit}$, 
  \textbf{Lei Liang}$^{\heartsuit\clubsuit}$, \\ 
  \textbf{Zhiqiang Zhang}$^{\heartsuit\clubsuit}$, 
  \textbf{Jun Zhou}$^{\heartsuit\clubsuit\dagger}$,
\textbf{Huajun Chen}$^{\spadesuit\heartsuit}$,
  \textbf{Ningyu Zhang}$^{\spadesuit\heartsuit}$\thanks{$\quad$ Corresponding Author.} \\
  $^\spadesuit$Zhejiang University    $^\clubsuit$Ant Group \\
  $^\heartsuit$Zhejiang University - Ant Group Joint Laboratory of Knowledge Graph \\
  \texttt{\{zhangjintian,zhangningyu\}@zju.edu.cn, jun.zhoujun@antgroup.com} \\
  \raisebox{-\mydepth}{\includegraphics[height=1.3\myheight]{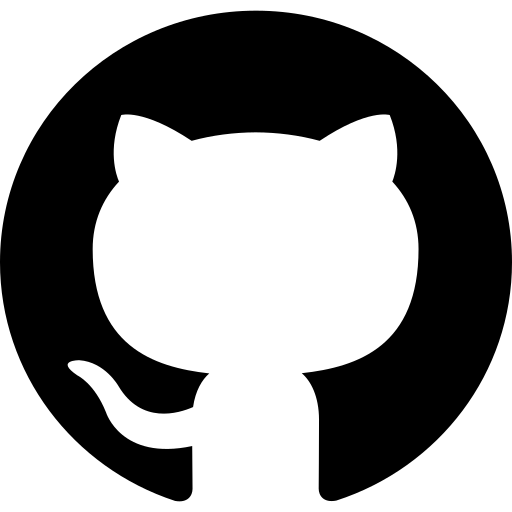}}
\textbf{\url{https://github.com/zjunlp/OneGen}}
}

\begin{document}

\maketitle

\begin{abstract}
Despite the recent advancements in Large Language Models (LLMs), which have significantly enhanced the generative capabilities for various NLP tasks, LLMs still face limitations in directly handling retrieval tasks. However, many practical applications demand the seamless integration of both retrieval and generation. This paper introduces a novel and efficient \textbf{One}-pass \textbf{Gen}eration and retrieval framework (\textbf{\ours}), designed to improve LLMs' performance on tasks that require both generation and retrieval. The proposed framework bridges the traditionally separate training approaches for generation and retrieval by incorporating retrieval tokens generated autoregressively. This enables a single LLM to handle both tasks simultaneously in a unified forward pass. We conduct experiments on two distinct types of composite tasks, RAG and Entity Linking, to validate the pluggability, effectiveness, and efficiency of \ours~in training and inference. Furthermore, our results show that integrating generation and retrieval within the same context preserves the generative capabilities of LLMs while improving retrieval performance. To the best of our knowledge, \ours~is the first to enable LLMs to conduct vector retrieval during the generation.
\end{abstract}

\section{Introduction}

In the era of Large Language Models (LLMs), many Natural Language Processing (NLP) tasks can be reduced to generation, allowing them to be addressed by a single LLM ~\citep{DBLP:journals/corr/abs-2303-18223,DBLP:conf/emnlp/QinZ0CYY23,DBLP:journals/corr/abs-2303-08774,DBLP:journals/corr/abs-2406-12793}. 
While LLMs excel in language generation, they still suffer from hallucinations (e.g., factual inaccuracies), stemming from their exclusive reliance on the parametric knowledge they contain~\citep{DBLP:journals/corr/abs-2309-01219,DBLP:conf/emnlp/YaoWT0LDC023,DBLP:journals/corr/abs-2401-01313}.  

\begin{figure}[!t] 
    \centering
    \scalebox{1}{
    \includegraphics[width=1\linewidth]{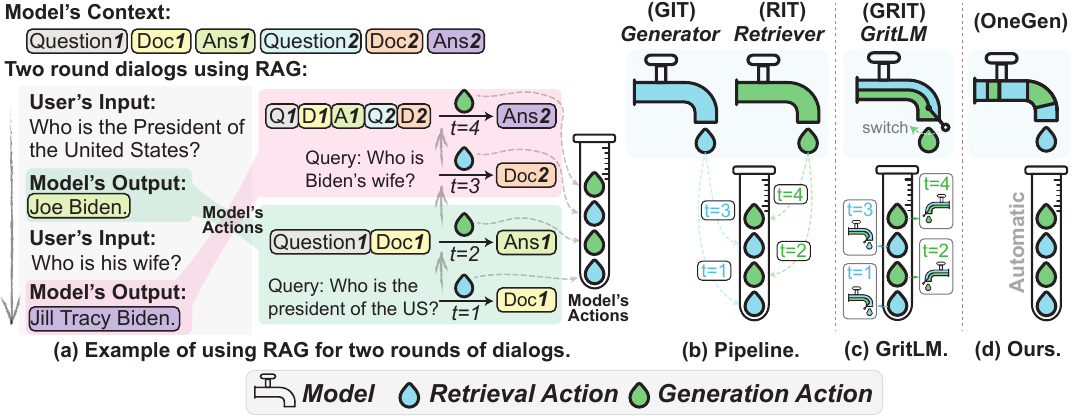} 
    }
    \caption{
    Comparison of Three Methods for RAG Task.
    (a) Two round dialogs using RAG  (Retrieve and Generate twice each).
    (b) Pipeline approach requiring the deployment of two separate models for retrieval and generation, 
    (c) GritLM~\citep{arxiv2024_GritLM} utilizing a single model with a switching mechanism to integrate retrieval and generation, 
    (d) OneGen (Ours) performing both functions automatically in the same model and the same context.
    }
    \vspace{-3mm}
    \label{fig:intro}
\end{figure}

One promising approach is Retrieval-Augmented Generation (RAG)~\citep{nips20_rag, emnlp23_active_rag, iclr24_selfrag,DBLP:journals/corr/abs-2405-14431,DBLP:journals/corr/abs-2312-10997}, which augments the input by retrieving relevant passages based on the query either before or during generation.
Other methods~\citep{arxiv24_EntGPT, acl24_chatkbqa} anchor LLM generation to an external knowledge base through Entity Linking (EL) during or after generation. These systems typically rely on a \textit{retriever} at various stages of generation. 
However, due to the separate training paradigms for generation and retrieval, most prior work \cite{arxiv2024_GritLM} employs a separate model for text embedding.
However, this pipeline approach has several drawbacks:
\textit{i)} Deploying and maintaining two separate models introduces additional hardware overhead and increases maintenance costs.
\textit{ii)} The separation of models creates two distinct representational spaces, limiting interaction between the retriever and generator (e.g., LLM) to text (i.e., query). As a result, whether the query is generated by the LLM or input directly by the user, it requires an additional forward pass through the retriever, increasing inference computational costs.
\textit{iii)} In multi-turn dialogues, as illustrated in Figure~\ref{fig:intro}(a), query rewriting is required for follow-up questions like ``Who is his wife?''. 
This rewriting adds inference overhead and risks error propagation if inaccurate.
\textit{iv)} Additionally, the pipeline approach is difficult to optimize end-to-end and requires large amounts of training data, while end-to-end optimization has been shown to yield significant benefits~\citep{iclr24_RA-DIT}.

Our work introduces an efficient \textbf{One}-pass unified \textbf{Gen}eration and retrieval (\textbf{\ours}) framework to enable an arbitrary LLM to generate and retrieve in one single forward pass. 
Inspired by the latest success in LLM for text embedding~\citep{arxiv24_e5_mistral}, we expand the original vocabulary by adding special tokens  (i.e. \textit{retrieval tokens}) and allocate the retrieval task to retrieval tokens generated in an autoregressive manner. 
During training, \textit{retrieval tokens} only participate in representation fine-tuning through contrastive learning~\citep{arxiv18_infonce,uai09_BPR}, whereas other output tokens are trained using language model objectives. 
At inference time, we use \textit{retrieval tokens} for efficient retrieving on demand. 

Unlike previous pipeline approaches, which require at least two models for retrieval and generation (as shown in Figure~\ref{fig:intro}(b)), \ours~unifies both tasks into a single model, eliminating the need for a separate retriever.
\cite{arxiv2024_GritLM} presents Generative Representational Instruction Tuning (GRIT), which aligns with this approach by training one LLM to handle both generative and embedding tasks through different prompts and attention mechanisms, as depicted by the \textit{``switch''} in Figure~\ref{fig:intro}(c). However, GRIT still necessitates independent forward passes for generation and retrieval tasks, reducing efficiency for tasks that intertwine generation and retrieval.

We evaluate the effectiveness of our method on two main tasks that require both generation and retrieval: RAG (including single-hop QA which needs single-retrieval and multi-hop QA which needs multi-retrieval) and Entity Linking (EL). 
Empirical results show \ours~outperforms the previous pipeline solutions as well as GRIT where applicable. 
Specifically, \ours~achieves +1.5pt improvement on average with four Single-hop QA datasets on top of Self-RAG~\citep{iclr24_selfrag}, +3.3pt F1 on average with two Multi-hop QA datasets under three different 7B-based LLMs, and +3.2pt accuracy on average with 6 out-of-domain entity linking datasets, with less training data. 
Moreover, further analysis demonstrates \ours~can enhance retrieval capability when jointly trained, with no sacrifice in generation capability. 
In addition, we demonstrate superior inference speed and memory consumption of \ours~compared with other LLM alternatives, particularly as retrieval frequency increases. 
In summary, our work makes the following \textbf{contributions}:

\textit{i)} We propose a training-efficiency, inference-efficiency, and pluggable framework OneGen that is particularly suitable for tasks interleaved with generation and retrieval. 
\textit{ii)} Our model, fine-tuned on less training data, demonstrates superior performance on six RAG datasets and six entity linking datasets on average.
\textit{iii)} We demonstrate the efficiency of \ours~at inference, highlighting a significant speed improvement as the length of query increases or retrieval frequency increases, compared to other LLM alternatives.
\textit{iv)} From the perspective of methodology, \ours~is an extension of Generative Instruction Tuning (GIT) and Representative Instruction Tuning (RIT) (as shown in Figure ~\ref{fig:intro}(b)).
\textit{v)} We contribute to communities by releasing our dataset as well as code.
\section{Preliminaries and Related Works}
%
Most text-based tasks can be reduced to generation, retrieval, or combination of the two.
We first introduce several hybrid tasks and their common solutions in \S~\ref{sec:pre:generation_retrieval}.
Then, we introduce the three roles of tokens in LLMs in \S~\ref{sec:pre:role}.
Finally, we further explain the motivation of our method in \S~\ref{sec:pre:motivation}.

\vspace{-1mm}
\subsection{Generation \& Retrieval}
\label{sec:pre:generation_retrieval}

\begin{wraptable}{r}{6.1cm}
\centering
\scalebox{0.75}{
\begin{tabular}{lccc}
\toprule
Task  & Input & $t=1$& $t=2$\\ \midrule
Generation  & $u$ & $y=G(u)$ & $-$  \\
Retrieval & ${u, \mathcal{K}}$ &  $\varepsilon = R(u, \mathcal{K}) $ & $-$  \\
$R\rightarrow G$ & ${u, \mathcal{K}}$ &  $\varepsilon = R(u, \mathcal{K}) $ &  $y=G(u, \varepsilon)$ \\ 
$G\rightarrow R$ & ${u, \mathcal{K}}$  & $y=G(u)$ & $\varepsilon = R(y, \mathcal{K})$\\ \cmidrule{1-4}
Unified &   ${u, \mathcal{K}}$ & \multicolumn{2}{c}{${y, \varepsilon} = OneGen(u, \mathcal{K})$} \\ 
\bottomrule
\end{tabular}
}
\caption{Comparison of different tasks and our unified solution.}
\label{table:pre:task}
\vspace{-1.5mm}
\end{wraptable}
For NLP problem related to generation or retrieval, a user input or a query $u=\{u_1,...,u_{n}\}$ and optionally document corpus $\mathcal{K}=\{d_i\}^{\left \| \mathcal{K} \right \|}_{i=1}$ are given (e.g., wiki articles), the end goal of the task is to generate sequence output $y=\{y_1,...,y_{m}\}$ or the most relevant documents $\epsilon$ from $\mathcal{K}$ with respect to $u$ or both.
We also assume that each $d_i\in\varepsilon$ is aligned to a subsequence or a whole sequence of tokens in $u$.
We summarize the steps and typical input, output for generation, retrieval, and two hybrid tasks in Table~\ref{table:pre:task}.

$\textbf{R}\rightarrow \textbf{G Task}$ leverages retrieval results to drive generation.
In the simplest format, a dense retrieval model (e.g., a dense passage retriever, DPR) is used to retrieve a collection of relevant documents $\varepsilon$ given user input $u$ at t=1;
$\varepsilon$ are then used as additional context when generating the target sequence using a generator (e.g. LLM) at t=2.
Retrieval Argumented Generation (RAG) is a classic example of $R\rightarrow G$ task.
Though there are some efforts in training the two model end-to-end predate the LLM era~\citep{nips20_rag}, most recent work use an off-the-shelf-retriever such as Contriever~\citep{tmlr22_contriever}, BM25, or search engine~\citep{emnlp23_active_rag}.
Furthermore, this task can involve multiple iterations of retrieval and generation, such as in multi-hop reasoning datasets like 2WIKI~\citep{coling20_2wiki} and HotpotQA~\citep{emnlp18_hotpotqa}.

$\textbf{G}\rightarrow \textbf{R Task}$ outputs retrieved documents relevant to user query in addition to generated content and are widely encountered in Information Retrieval (IR).
A prominent example task is Entity Linking (EL), which involves locating mentions and disambiguating these surface forms into entities in some Knowledge Base (KB).
Early EL methods~\citep{acl11_Hoffmann} treat EL as decomposed subtasks, such as Mention Detection (MD) and Entity Disambiguation (ED), and solve them in sequence.
More recent works manage to frame EL as an end-to-end task, such as sequence generation~\citep{iclr21_GENRE}, question answering~\citep{iclr22_entqa}, retrieve augmented generation~\citep{emnlp23_INSGENEL}, and sequence tagging problem~\citep{conll19_end2end_el,naacl22_refined}, which outperform the early pipeline approach.
For the generative EL paradigm, MD can be modeled as a generation task where entities in the original sentences are generated; ED is a typical retrieval task of retrieving the most relevant entity from the KB given a mention span.

\vspace{-1mm}
\subsection{Roles of tokens in LLMs}
\label{sec:pre:role}
A token $x_i$ is the basic unit processed by an LLM.
Token in the input of an LLM serves three different roles:
\textit{1) generating the next token}, noted as ${role}(x_i)=\texttt{GEN}$;
\textit{2) providing context information}, noted as ${role}(x_i)=\texttt{CTX}$;
and \textit{3) representing a sentence}, noted as ${role}(x_i)=\texttt{RET}$.
Recent works~\citep{arxiv24_e5_mistral,arxiv2024_GritLM} use the hidden state of the last token as the sentence representation.

\vspace{-1mm}
\subsection{Motivation}
\label{sec:pre:motivation}
Recent years have seen a rise in using LLMs to handle complex hybrid tasks, replacing traditional NLP model pipelines. 
Before LLMs, end-to-end approaches offered advantages for combining generation and retrieval tasks, reducing error propagation compared to pipelines and potentially improving efficiency with single-pass inference. 
However, earlier solutions are often task-specific and lack generalization across hybrid tasks. 
For instance, in generative EL, methods like constrained decoding~\citep{iclr21_GENRE} are used to retrieve entities efficiently. 
Our work addresses the absence of a unified LLM framework for hybrid tasks, stemming from separate training approaches for generation and retrieval tasks, which typically use distinct objectives and datasets.

\section{OneGen: One-Pass Generation and Retrieval for LLMs}
\vspace{-2mm}
\begin{figure*}[!t] 
    \centering
    \scalebox{0.9}{
    \includegraphics[width=1.0\linewidth]{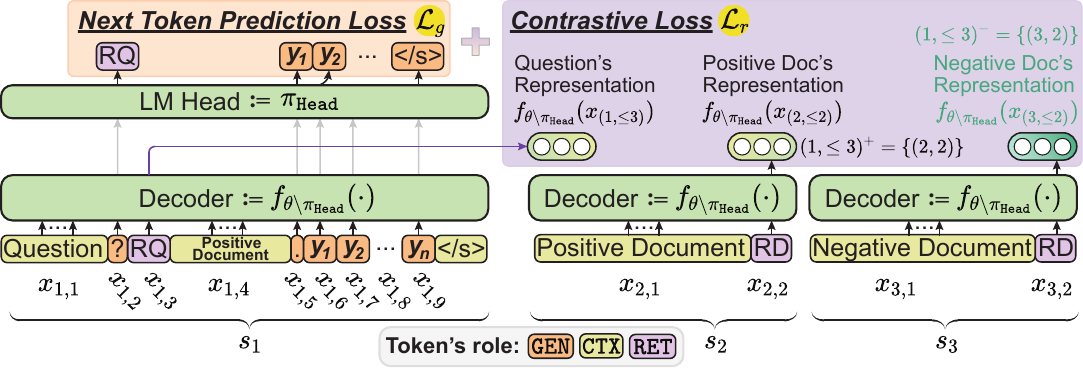} 
    }
    \vspace{-1mm}
    \caption{The training framework of unified \textbf{One}-pass \textbf{Gen}eration and retrieval (\textbf{OneGen}), illustrated using RAG. Detailed training process for other tasks can be found in Figure~\ref{fig:app:training} of Appendix.}
    \label{fig:method}
    \vspace{-3mm}
\end{figure*} 
We introduce a \textbf{One}-pass \textbf{Gen}eration and retrieval framework (\textbf{OneGen}) for fine-tuning LLMs on generation, retrieval, or hybrid tasks, as shown in Figure~\ref{fig:method}.
Our core idea is to integrate generation and retrieval to the same context by allocating the retrieval task to \textit{retrieval tokens} generated in an autoregressive manner, thus enabling LLM to perform both tasks in a single forward pass.
\vspace{-1.5mm}
\subsection{Overview}
\vspace{-0.5mm}
\textbf{Notation.}
To ensure clarity and precision in our subsequent discussions, we standardize the notation used in Table~\ref{table:pre:task}.
Define the dataset $\mathcal{D}=\{s_i\}_{i=1}^{\left | \mathcal{D} \right |}$, which consists of $\left | \mathcal{D} \right |$ sentences $s$ of varying lengths, with each sentence $s=\{x_i\}_{i=0}^{\left | s \right |}$ comprising $\left | s \right |$ tokens $x$.
Let $x_{i,j}$ denote the $j$-th token of the $i$-th sentence in the dataset $\mathcal{D}$, and define $x_{i,\le j}$ as $\{x_{i,1},x_{i,2},\ldots,x_{i,j}\}$.
We can distinguish the symbols $u$, $y$, and $d$ defined in Table~\ref{table:pre:task} based on the role of tokens $x$ within the sentence $s$.
Specifically, $y$ corresponds to the segment of the $s$ where $role(x)=\texttt{GEN}$ , $u$ corresponds to the segment where $role(x)=\texttt{CTX}$, and if all tokens $x$ in a sentence $s$ have $role(x)=\texttt{CTX}$, then it corresponds to $d$.
Given the instruction dataset $\mathcal{I}$, where $s=\{u,y\}\in\mathcal{I}$, we have $\mathcal{D}=\mathcal{I}\cup\mathcal{K}$.

\textbf{Design.}
Retrieval requires encoding both the query and the document within the same representational space.
\textit{Our core idea is to incorporate query encoding into the generation process.}
Thus we use the same LLM for encoding both the query and the document, without altering the model structure, such as the attention mechanism, unlike the approach taken by GritLM.
Specifically, for query encoding, we introduce a special token $x_i=\texttt{[RQ]}$, where $role(x_i)=\texttt{RET}$.
This token is generated by the LLM and used as input to represent the query.
However, assigning $role(x_i)=\texttt{RET}$ prevents the generation of the next token $x_{i+1}$ if $role(x_{i+1})=\texttt{GEN}$.
To address this, we also introduce a \texttt{<CON>} token during data reconstruction, ensuring the continuation of the generation process.

\textbf{Inference.}
At inference time, the documents to be retrieved are encoded offline by the trained LLM using the template ``\texttt{\{document\}[RD]}'', where $role(\texttt{[RD]})=\texttt{RET}$.
Then the trained LLM autoregressively generates tokens based on the user's input until it encounters the \texttt{[RQ]} token.
The logits corresponding to the \texttt{[RQ]} token are then used for retrieval.
Depending on the task requirements, the retrieved content may be concatenated with the context, potentially along with the \texttt{<CON>} token, before continuing with the inference until the generation is complete.
\vspace{-1.5mm}
\subsection{Train}
\label{sec:method:train}
\vspace{-0.5mm}
\textbf{Data Reconstruction.}
We augment the standard generation output with retrieval tokens wherever retrieval is needed.
This makes our framework easily pluggable to existing methods.
Generally, we insert \texttt{[RQ]} to sentence $s$ for query representation.
In particular, if the query span is explicit, we add optional tokens \texttt{<LOC>} and \texttt{</LOC>} to assist in locating the position of the query.
The augmented sequence is $s=\{x_{\le i}, \texttt{<LOC>}, x_{i+1}, \ldots, x_{j}, \texttt{</LOC>},\texttt{[RQ]},\texttt{<CON>},x_{(>j)}\}$.
The token $\texttt{<CON>}$ enables LLMs to generate continuously and it must be included if and only if $role(x_{j+1})=\texttt{GEN}$, i.e., generation is required after retrieval.
For each document $x\in\mathcal{K}$, \texttt{[RD]} is usually appended to the end of the document to represent the document.
In particular, we can add \texttt{[RD]} for each end of the sentence in a document $x\in\mathcal{K}$ to get the fine-grained representation.
Figure~\ref{app:fig:rg:self_rag:doc:case} and Figure~\ref{app:fig:rg:multi_hop:doc:case} in appendix show two different examples for reconstructed document .

\textbf{Training Objective.}
The optimization is only performed for tokens $x_i\in s$ where $role(x_i)\in\{\texttt{GEN}, \texttt{RET}\}$.
A simple application of \ours~in the RAG task is illustrated in Figure~\ref{fig:method}.
Note that, $role(x_i)=\texttt{RET}$ iff $x_i\in\{\texttt{[RQ]},\texttt{[RD]}\}$ (highlight in purple in Figure~\ref{fig:method}).
For tokens where $role(x_i)=\texttt{GEN}$ (highlight in orange), optimization employs $\mathcal{L}_g$:
\vspace{-1.5mm}
$$
\mathcal{L}_{g}
=
\frac{1}{\left | \mathcal{D} \right |} 
{\sum_{i=1}^{\left | \mathcal{D} \right |}}
{\sum_{j=1}^{\left | s_i \right |}}
\ell_{g}(
f_{\theta\setminus\pi_{\texttt{Head}}}(x_{(i,{\le j})}), \pi_{\texttt{Head}})
)
\cdot
\mathbbm{1}_{g}(x_{i,j}).
\vspace{-1.5mm}
$$
Here, $\theta$ is the LLM parameter. $\pi_{\texttt{Head}}\in\mathbbm{R}^{N\times d}$ denotes the expanded vocabulary (i.e., LM Head), consists of $N$ $d$-dimensional vectors.
$f_{\theta\setminus\pi_{\texttt{Head}}}(x_{(i,{\le j})})\in\mathbbm{R}^d$ denotes a $d$-dimensional vector generated by the LLM without LM Head from processing the first to the $j$-th token.
$\ell_g$ typically represents the cross-entropy loss function, and $\mathbbm{1}_g(x_{i,j})$ is an indicator function, where $\mathbbm{1}_g(x_{i,j})=1$ iff $role(x_{i,j})=\texttt{GEN}$; otherwise, it is 0.
For tokens where $role(x_i)=\texttt{RET}$, optimization employs $\mathcal{L}_r$:
\vspace{-2mm}
$$
\mathcal{L}_{r}
=
\frac{1}{\left | \mathcal{D} \right |} 
{\sum_{i=1}^{\left | \mathcal{D} \right |}}
{\sum_{j=1}^{\left | s_i \right |}}
\ell_{r}\left (
f_{\theta\setminus\pi_{\texttt{Head}}}(x_{(i,{\le j})})\textbf{,}~ 
f_{\theta\setminus\pi_{\texttt{Head}}}(x_{(i,{\le j})^{+}})\textbf{,}~
f_{\theta\setminus\pi_{\texttt{Head}}}(x_{(i,{\le j})^{-}})
\right )
\cdot\mathbbm{1}_{r}(x_{i,j}).
\vspace{-2mm}
$$
Here, $\ell_r$ is the contrastive loss (e.g., InfoNCE~\citep{arxiv18_infonce}),  with $(i,\le j)^-$ and $(i,\le j)^+$ representing the sets of indices for negative and positive samples about sequence $x_{i,\le j}$, respectively.
The example in Figure~\ref{fig:method} illustrates this concept clearly and effectively.
$\mathbbm{1}_r(x_{i,j})$ is an indicator function, where $\mathbbm{1}_r(x_{i,j})=1$ iff $role(x_{i,j})=\texttt{RET}$; otherwise, it is 0.
Finally, combining the two parts of loss by weighted addition gives the final optimization goal: 
$\mathcal{L}=\lambda_g\mathcal{L}_g+\lambda_r\mathcal{L}_r$, where $\lambda_g$ and $\lambda_r$ are hyperparameters.
For a comprehensive overview of the detailed training procedures employed in other tasks, kindly refer to Figures~\ref{fig:app:training} in Appendix.

\textbf{Optimization.}
We use the standard Cross Entropy to optimize the loss function $\ell_{g}$. 
For the loss function $\ell_{r}$, prior work~\citep{arxiv2024_GritLM} often utilizes the InfoNCE, which requires GradCache~\citep{acl21_gradcache} to handle large batch sizes on low-memory GPUs, adding overhead and limiting to one positive sample per batch with a carefully chosen temperature hyperparameter.
In contrast, we employ the hyperparameter-free BPR~\citep{uai09_BPR}, a pair-wise loss function. 
The $\ell_{r}$ loss is defined as:
$\ell_{r} = -\log~\sigma\left(\left \|f(x_{(i,\le j)})\right \|\cdot\left(\left \|f(x_{(i,\le j)^+_{k}})\right \| - \left \|f(x_{(i,\le j)^-_{k}})\right\|\right)\right)$,
where $\sigma(\cdot)$ is the sigmoid function, $\left \| \cdot\right \|$ is normalization, and $(i,\le j)^+_{k}$ and $(i,\le j)^-_{k}$ are randomly selected from $(i,\le j)^+$ and $(i,\le j)^-$ respectively.
Using BPR allows for Gradient Accumulation to support larger batch sizes and multiple positive samples per batch. Experimental results in Table~\ref{table:exp:ablation:gr:loss} show that BPR reduces negative impacts on generative tasks and significantly enhances retrieval tasks.

\subsection{Inference}
For the standalone tasks of \textit{Generation} and \textit{Retrieval}, \ours~aligns with prior work.
Here, we exclusively address the hybrid task of Generation and Retrieval.
The inference process includes two primary steps:
\textbf{1) Cache document embedding.}
To facilitate efficient retrieval, we cache our document embedding after training is done, similar with prior work.
Specifically, we append \texttt{[RD]} at the end of each document $x\in\mathcal{K}$ and use $f_{\theta\setminus\pi_\texttt{Head}}$ to encoding them.
We process all the documents in batch and the collection of these representations is stored in $Emb_{doc}$ where $Emb_{doc}\in\mathbbm{R}^{\left | \mathcal{K} \right |\times d}$.
\textbf{2) Task generation.}
Given an instruction, \textit{2.1)} the LLM begins autoregressively generate the next token until $role(x_i)=\texttt{RET}$ or $x_i\in\texttt{Terminator}$ (e.g., \texttt{</s>} in Llama2).
If $role(x_i)=\texttt{RET}$, then $f_{\theta\setminus\pi_\texttt{Head}}(x_{\le i})$ is used to retrieve from $Emb_{doc}$ to obtain the set of most relevant documents $d_{r}\subset\mathcal{K}$.
Here we use cosine similarity.
\textit{2.2) How to precede to generate the next token $x_{i+1}$?}
Since $role(x_i)\ne\texttt{GEN}$, the probability distribution $P(x_{i+1}|x_{\le i})$ is not intrinsically modeled by LLM, thus, should be provided by user.
For the multi-turn dialogue system, the user provides the token that initiates a new round of dialogue rather than the LLM.
Specifically, for the $R\rightarrow G$ task, $P(x_{i+1}|x_{\le i})=d_{r}$ (i.e., directly concatenating the retrieved document).
For $R\rightarrow G$ task, we specify $P(x_{i+1}|x_{\le i})=\texttt{[CON]}$.
\textit{2.3)} Finally, the LLMs continue autoregressive prediction, repeating \textit{step 2.1}.
Detailed pseudocode and description can be found in Appendix~\ref{app:rg_single:inference} and Appendix~\ref{app:gr:inference}.

\textbf{Difference from Related Works.}
GritLM, GENRE~\citep{iclr21_GENRE}, and RICHES~\citep{arxiv_deepmind_riches} can all perform specific hybrid retrieval and generation tasks using a single model. 
GritLM with causal attention first generates a query, then re-encodes it with bidirectional attention for retrieval in continuous space (i.e., using vector).  
GENRE and RICHES generate a query explicitly and use it to constrain decoding, enabling retrieval during generation in discrete space (i.e., using tokens) with just one forward pass. 
In contrast, \ours~performs retrieval in continuous space during generation, avoiding the need for two forward passes.
Kindly refer to related work in Appendix~\ref{app:related_work} for details.
\section{Experiments}
\begin{table*}[!t]
\centering
\small
\scalebox{0.8}{
\begin{tabular}{cccclccccc}
\toprule
\multirow{3}{*}{\textbf{LLMs}} & \multicolumn{3}{c}{\textbf{Retriever}}           &  & \multicolumn{4}{c}{\textbf{Dataset}}                                   & \multirow{3}{*}{\textbf{AVG.}} \\ \cmidrule{2-4} \cmidrule{6-9}
                      & Name       & Dataset Name & Dataset Size      &  & PopQA         & TQA           & Pub           & ARC           &                      \\ \midrule
Toolformer~\citep{nips23_Toolformer}            & Contriever & MS MARCO & $1\times 10^{6}$ & & - & 48.8 & - & - & - \\
Llama2$_{\texttt{7B}}$~\citep{arXiv2023_LLaMA}             & Contriever & MS MARCO  & $1\times 10^{6}$ &  & 38.2 & 42.5 & 30.0 & 48.0 & 39.7 \\
Alpaca$_{\texttt{7B}}$~\citep{nips23_Alpaca}             & Contriever & MS MARCO  & $1\times 10^{6}$             &  & 46.7 & 64.1 & 40.2 & 48.0 & 49.8 \\
SAIL$_{\texttt{7B}}$~\citep{arxiv23_SAIL} & Contriever & MS MARCO & $1\times 10^{6}$ &  & - & - & 69.2 & 48.4 & - \\
Llama2-FT$_{\texttt{7B}}$~\citep{arXiv2023_LLaMA} & Contriever & MS MARCO & $1\times 10^{6}$ & & 48.7 & 57.3 & 64.3 & 65.8 & 59.0 \\
Mistral$_{\texttt{7B}}$~\citep{arXiv2023_Mistral} & Contriever & MS MARCO & $1\times 10^{6}$ &  & {23.2} & {49.3} & 52.0 & 39.0 & 40.9 \\ 
GritLM$_{\texttt{7B}}$~\citep{arxiv2024_GritLM} & GritLM$_{\texttt{7B}}$ & E5S(w/ TQA) & $2\times 10^{6}$ &  & \textbf{58.0} & \textbf{66.5} & 49.7 & 24.5 & 49.7 \\ \midrule
Self-RAG$_{\texttt{7B}}$~\citep{iclr24_selfrag} & Contriever & MS MARCO  & $1\times 10^{6}$ & & {\underline{52.5}} & 65.0 & {\underline{72.2}} & {\underline{67.3}} & {\underline{64.3}} \\
Self-RAG$_{\texttt{7B}}$~(+\ours) & \textit{Self} & \textit{Sampled} & \textbf{$\bf 6\times 10^{4}$} & & {\underline{52.5}} & {\underline{65.7}} & \textbf{75.1} & \textbf{70.1} & \textbf{65.8} \\ \bottomrule
\end{tabular}
}
\vspace{-1mm}
\caption{
Performance comparison across different datasets.
``TQA'' means TriviaQA, ``Pub'' means PublicHealth.
The best and second-best results are indicated in bold and underlined.
The complete table is shown in Table~\ref{table:exp_rag_all} of appendix.
The details about Self-RAG are shown in appendix~\ref{app:rg_single:self_rag}.
}
\label{table:exp_rag}
\vspace{-1.5mm}

\end{table*}
\subsection{Experimental Settings}
We conduct extensive experiments across three different settings to validate \ours's \textit{effectiveness}, \textit{efficiency in training and inference}, and \textit{pluggability}.
We train and evaluate a model for each setting independently.
Specifically, the Single-hop QA involves one round of $R\rightarrow G$, Multi-hop QA entails multiple $R\rightarrow G$ executions, and Entity Linking involves multiple $G\rightarrow R$ executions.
Due to page constraints, we provide a concise overview of the \textit{workflow}, \textit{baseline}, \textit{training data}, \textit{training backbones}, \textit{evaluation datasets}, and \textit{evaluation metrics} for each setting in order.
Detailed information for each following settings can be found in the Appendix~\ref{app:rg_single:experiments},~\ref{app:rg_multi:experiments}, and~\ref{app:gr:experiments}, respectively.

$\textbf{R} \rightarrow \textbf{G Task: RAG for Single-hop QA.}$
We apply \ours~directly to the Self-RAG~\citep{iclr24_selfrag} method (i.e., using Self-RAG as workflow), which incorporates adaptive retrieval and self-assessment.
\ours~enhances it by enabling self-retrieval.
The baselines we used are listed in Table~\ref{table:exp_rag}.
Llama2-7B-chat serves as the backbone of ours and Self-RAG.
The training data is derived from the Self-RAG training dataset, modified according to \S~\ref{sec:method:train}, comprising 150k instances.
We construct positive and negative samples using heuristic rules for instances containing the \texttt{[RQ]} token.
These samples are sourced from \textit{wiki}, comprising 60k instances.
We evaluate four datasets:
PubHealth~\citep{arxiv23_PubHealth}, ARC-Challenge~\citep{arxiv18_ARC}, PopQA~\citep{acl23_PopQA}, and TriviaQA~\citep{acl17_TriviaQA}.
For the first two, accuracy serves as the evaluation metric.
For the others, evaluation is based on whether the model's output contained the ground truth.
All evaluations are consistent with Self-RAG to ensure fair comparison.

$\textbf{R} \rightarrow \textbf{G Task: RAG for Multi-hop QA.}$
The following text sequence effectively demonstrates the workflow:``
\textcolor{CTX}{{\{Instruction\} Are Alan Turing and Newton from the same country?}}
\textcolor{GEN}{{What is Alan Turing's country?}}
\textcolor{RET}{\texttt{[RQ]}}
\textcolor{CTX}{{\{document 1\}}}
\textcolor{GEN}{{England. What is Newton's country?}}
\textcolor{RET}{\texttt{[RQ]}}
\textcolor{CTX}{{\{document 2\}}}
\textcolor{GEN}{{England. Therefore, yes.}}''.
This workflow combines CoT and RAG, where the gray tokens represent the role of \texttt{CTX}, the purple tokens represent \texttt{RET}, and the orange tokens represent \texttt{GEN}.
The baseline uses the Contriever for retrieval, and the generator in the baseline is trained on the constructed data, with the optimization objective solely being $\mathcal{L}_g$.
We sample 10\% of the training dataset from HotpotQA and 2WIKI, using Qwen2-72B~\citep{arxiv23_qwen} for data construction.
Heuristics rules are also employed to label the positive and negative samples.
Evaluation is conducted using the validation set of HotpotQA and 2WIKI, as the ground truth of test sets is unavailable.
We use EM and F1 to evaluate the model's generative capability and Recall@1 for its retrieval capability.

$\textbf{G} \rightarrow \textbf{R Task: Entity Linking.}$
The following text sequence effectively shows the workflow:``
\textcolor{CTX}{\{Instruction\} Steve Jobs founded Apple Inc.}
\textcolor{GEN}{\texttt{<LOC>}Steve Jobs\texttt{</LOC>}}
\textcolor{RET}{\texttt{[RQ]}}
\textcolor{GEN}{\texttt{<CON>} founded \texttt{<LOC>}Apple Inc\texttt{</LOC>}}
\textcolor{RET}{\texttt{[RQ]}}
\textcolor{GEN}{\texttt{<CON>}.}''.
The baselines we used are listed in Table~\ref{table:exp_el}.
We employ the Wikipedia (totaling 6M documents, yet randomly sampling only 60K without careful selection) and AIDA~\citep{emnlp11_aida} datasets, applying data augmentation to each sample in Wikipedia following established methods from previous studies~\citep{emnlp21_wikihyperlink}.
We adopt heuristic rules to label each mention that includes an entity ID with both positive and negative documents, as detailed in the Appendix~\ref{app:gr:training}.
Llama2-7B-chat also serves as the backbone.
We utilize the ELEVANT~\citep{emnlp22_ELEVANT} for evaluation across seven datasets listed in Table~\ref{table:exp_el}, using Micro F1 to evaluate in-KB entities.
The same datasets and metrics are applied to MD task.
For ED task, following the ChatEL~\citep{coling24_ChatEL}, we evaluate nine datasets, maintaining the use of the Micro F1.

\subsection{Main Results}
\label{sec:exp:main}
\noindent
\vspace{-2pt}
\begin{figure}[t]
\begin{minipage}{0.66\textwidth} 
    \small
    \scalebox{0.75}{
    \begin{tabular}{cccccccc}
    \toprule
    \multicolumn{1}{l}{} & \multicolumn{1}{l}{} & \multicolumn{4}{c}{\textbf{Generation Performance}} & \multicolumn{2}{c}{\textbf{Retrieval Performance}} \\  \cmidrule(lr){3-6} \cmidrule(lr){7-8} 
    \multicolumn{1}{l}{} & \multicolumn{1}{l}{} & \multicolumn{2}{c}{HotpotQA} & \multicolumn{2}{c}{2WIKI} & HotpotQA & 2WIKI \\  \cmidrule(lr){3-4} \cmidrule(lr){5-6} \cmidrule(lr){7-7} \cmidrule(lr){8-8}
    \textbf{BackBone} & \textbf{Retriever} & EM & F1 & EM & F1 & Recall@1 & Recall@1 \\  \midrule
     & \multicolumn{1}{c|}{Contriever} & 52.83 & 65.64 & 70.02 & \multicolumn{1}{c|}{74.35} & 73.76 & 68.75 \\
    \multirow{-2}{*}{Llama2-7B} & \multicolumn{1}{c|}{\cellcolor[HTML]{EFEFEF}\textit{self}} & \cellcolor[HTML]{EFEFEF}\textbf{54.82} & \cellcolor[HTML]{EFEFEF}\textbf{67.93} & \cellcolor[HTML]{EFEFEF}\textbf{75.02} & \multicolumn{1}{c|}{\cellcolor[HTML]{EFEFEF}\textbf{78.86}} & \cellcolor[HTML]{EFEFEF}\textbf{75.90} & \cellcolor[HTML]{EFEFEF}\textbf{69.79} \\  \midrule
     & \multicolumn{1}{c|}{Contriever} & 53.72 & 66.46 & 70.92 & \multicolumn{1}{c|}{75.29} & 69.79 & 66.80 \\
    \multirow{-2}{*}{Llama3.1-7B} & \multicolumn{1}{c|}{\cellcolor[HTML]{EFEFEF}\textit{self}} & \cellcolor[HTML]{EFEFEF}\textbf{55.38} & \cellcolor[HTML]{EFEFEF}\textbf{68.35} & \cellcolor[HTML]{EFEFEF}\textbf{75.88} & \multicolumn{1}{c|}{\cellcolor[HTML]{EFEFEF}\textbf{79.60}} & \cellcolor[HTML]{EFEFEF}\textbf{72.55} & \cellcolor[HTML]{EFEFEF}\textbf{68.98} \\  \midrule
     & \multicolumn{1}{c|}{Contriever} & 48.55 & \textbf{61.02} & 68.32 & \multicolumn{1}{c|}{72.66} & 72.41 & 67.70 \\
    \multirow{-2}{*}{Qwen2-1.5B} & \multicolumn{1}{c|}{\cellcolor[HTML]{EFEFEF}\textit{self}} & \cellcolor[HTML]{EFEFEF}\textbf{48.75} & \cellcolor[HTML]{EFEFEF}60.98 & \cellcolor[HTML]{EFEFEF}\textbf{73.84} & \multicolumn{1}{c|}{\cellcolor[HTML]{EFEFEF}\textbf{77.44}} & \cellcolor[HTML]{EFEFEF}\textbf{72.70} & \cellcolor[HTML]{EFEFEF}\textbf{69.27} \\  \midrule
     & \multicolumn{1}{c|}{Contriever} & 53.32 & 66.22 & 70.80 & \multicolumn{1}{c|}{74.86} & 74.15 & 69.01 \\
    \multirow{-2}{*}{Qwen2-7B} & \multicolumn{1}{c|}{\cellcolor[HTML]{EFEFEF}\textit{self}} & \cellcolor[HTML]{EFEFEF}\textbf{55.12} & \cellcolor[HTML]{EFEFEF}\textbf{67.60} & \cellcolor[HTML]{EFEFEF}\textbf{76.17} & \multicolumn{1}{c|}{\cellcolor[HTML]{EFEFEF}\textbf{79.82}} & \cellcolor[HTML]{EFEFEF}\textbf{75.68} & \cellcolor[HTML]{EFEFEF}\textbf{69.96} \\ \bottomrule
    \end{tabular}
    }
    \vspace{-1.5mm}
    \captionof{table}{
    In RAG for Multi-Hop QA settings, performance comparison across different datasets using different LLMs.
    }
    \label{table:exp_cot_rag}
\end{minipage}\hfill 
\begin{minipage}{0.32\textwidth} 


    \vspace{-1mm}
    \centering
    \small
    \scalebox{0.9}{
    \begin{tabular}{lcc}
    \toprule
    \multicolumn{1}{c}{\multirow{3}{*}{\textbf{Task}}} & \multicolumn{2}{c}{\textbf{\begin{tabular}[c]{@{}c@{}}Loss Function\\ $(\mathcal{L}_r)$\end{tabular}}} \\ \cmidrule{2-3} 
    \multicolumn{1}{c}{}                               & InfoNCE           & BPR                    \\ \midrule
    EL (7 datasets)                                              & 61.8              & \textbf{64.0}          \\
    ED (9 datasets)                                              & 84.5              & \textbf{86.5}          \\
    MD (7 datasets)                                              & 70.7              & \textbf{71.5}          \\ \bottomrule
    \end{tabular}
    }
    \captionof{table}{Ablation study results of $\mathcal{L}_r$ on EL, ED, and Mention Detection (MD) tasks.
    The table reports average F1 scores for each task.}
    \label{table:exp:ablation:gr:loss}
    
\end{minipage}

\end{figure}

\vspace{-2mm}
\begin{table*}[]
\centering
\small
\scalebox{0.75}{
\begin{tabular}{lccccccccccc}
\toprule
\multirow{3}{*}{\textbf{Method}} & \multirow{3}{*}{\textbf{Cand. Size}} & \multirow{3}{*}{\textbf{Training Data$^\blacklozenge$}} & \textbf{In-domain}     &  & \multicolumn{6}{c}{\textbf{Out-of-domain}}                                                             & \multirow{3}{*}{\textbf{AVG.}} \\ \cmidrule{4-4} \cmidrule{6-11}
                       &                            &                                & AIDA          &  & OKE15         & OKE16         & REU           & MSN           & SPOT          & K50        &                      \\ \midrule
Neural EL$^\blacklozenge$ & $<30$ & AIDA & 76.3 &  & 60.6 & 53.8 & 44.0 & 56.5 & 19.5 & 38.2 & 49.8 \\
REL 2019$^\diamondsuit$ & $<30$ & - & 85.4 &  & {\underline{66.5}} & 57.7 & {\underline{53.0}} & \textbf{77.8} & \underline{24.9} & 54.0 & 59.9  \\
GENRE$^\blacklozenge$ & $<30$ & WIKI 6M+AIDA & {\underline{85.3}}  &  & 54.9 & 44.4 & 46.3 & 69.3  & 24.6 & 56.9 & 54.5 \\ 
ReFinED$^\blacklozenge$ & $<30$ & WIKI 6M+AIDA & \textbf{88.6} &  & \textbf{66.6} & {\underline{61.2}} & 49.8 & {\underline{74.7}} & 22.2 & {\underline{62.8}} & \underline{60.8}                 \\ \midrule
Llama2$_{\texttt{7B}}$~(+\ours)$^\blacklozenge$ & 1.25M & WIKI 60K+AIDA & 83.1 &  & 63.5 & \textbf{64.3} & \textbf{61.1} & 74.2 & {\textbf{28.8}} & \textbf{72.7} & \textbf{64.0} \\ \bottomrule
\end{tabular}
}
\caption{
EL task performance on in-domain and out-of-domain test sets.
The best value is in bold and the second best is underlined.
The `$\blacklozenge$' denotes end2end method, while the `$\diamondsuit$' denotes pipelines.
}
\label{table:exp_el}
\vspace{-1.5mm}
\end{table*}
\vspace{-1.5mm}
In Table~\ref{table:exp_rag}, Table~\ref{table:exp_el}, and Table~\ref{table:exp_cot_rag}, we report the performance of \ours~on three types of settings respectively, demonstrating that our method is both \textit{effective}, \textit{pluggable} and \textit{training-efficient}.

$\textbf{R} \rightarrow \textbf{G Task for Single-hop QA.}$
From Table~\ref{table:exp_rag}, we draw the following conclusions:
\textit{(1) \ours~demonstrates efficacy in $R\rightarrow G$ task, and joint training of retrieval and generation yields performance gains on the RAG task.}
The Self-RAG endows LLMs with self-assessment and adaptive retrieval, while \ours~adds self-retrieval.
Our method outperforms the original Self-RAG across all datasets, especially achieving improvements of 3.1pt on Pub dataset and 2.8pt on ARC dataset, validating the benefits of joint training, consistent with findings from RA-DIT~\citep{iclr24_RA-DIT} and GRIT~\citep{arxiv2024_GritLM}.
However, ours on PopQA and TQA datasets remains inferior to GritLM-7B.
We attribute this to using a larger retrieval dataset, E5S, which is twice the size of MS MARCO and 33 times larger than \ours. 
Additionally, E5S includes TQA training data and higher-quality data.
\textit{(2) \ours~is highly efficient in training, with instruction-finetuned LLMs showing strong retrieval capabilities with minimal additional tuning.}
It requires less and lower-quality retrieval data, achieving comparable performance with just 60K noisy samples and incomplete documents, without synthetic data.
This efficiency aligns with findings from previous works like PromptEoL~\citep{arxiv23_PromptEOL} and EchoEmbedding~\citep{arxiv24_EchoEmb}, which demonstrate excellent performance using prompt-based methods without further training.

$\textbf{R} \rightarrow \textbf{G Task for Multi-hop QA.}$
From Table~\ref{table:exp_cot_rag}, we additionally find that \textit{\ours~remains effective across multiple $R\rightarrow G$ settings and works well with various models and scales.} 
It consistently outperforms the baseline on most datasets, backbones, and metrics. 
Notably, on the 2WIKI, \ours~achieves an average improvement of 5.2pt in EM and 4.6pt in F1. 
Additionally, our evaluation of end-to-end retrieval performance indicates that \ours's performance on retrieval surpasses the baseline across all datasets and backbones.

$\textbf{G} \rightarrow \textbf{R Task.}$
From Table~\ref{table:exp_el}, we can draw the following conclusions:
\textit{(1) \ours~demonstrates effectiveness and strong generalization in the $G\rightarrow R$ task.}
It outperforms ReFinED in the F1 score by 3.2pt on average and surpasses the most competitive baselines on the OKE16, REU, and K50 datasets by 3.1pt, 8.1pt, and 10.1pt, respectively.
The lower performance on other datasets is attributed to insufficient training data, while the baselines utilized 6M data and we used only 1\% of this amount.
As INSGENEL~\citep{emnlp23_INSGENEL} has shown, increased training data can improve MD performance.
In \S~\ref{sec:exp:impacts}, we analyze the bottlenecks in datasets with poorer outcomes, which lie in MD.
Additionally, only 24\% of entities in our candidate set participated in the training process.
\textit{(2) \ours~is highly efficient in training,} requiring merely 1\% of data used in baselines.

\subsection{Analysis}
\subsubsection{Efficiency at Inference Time}
\begin{figure*}[] 
    \centering
    \scalebox{1.0}{
    \includegraphics[width=1\linewidth]{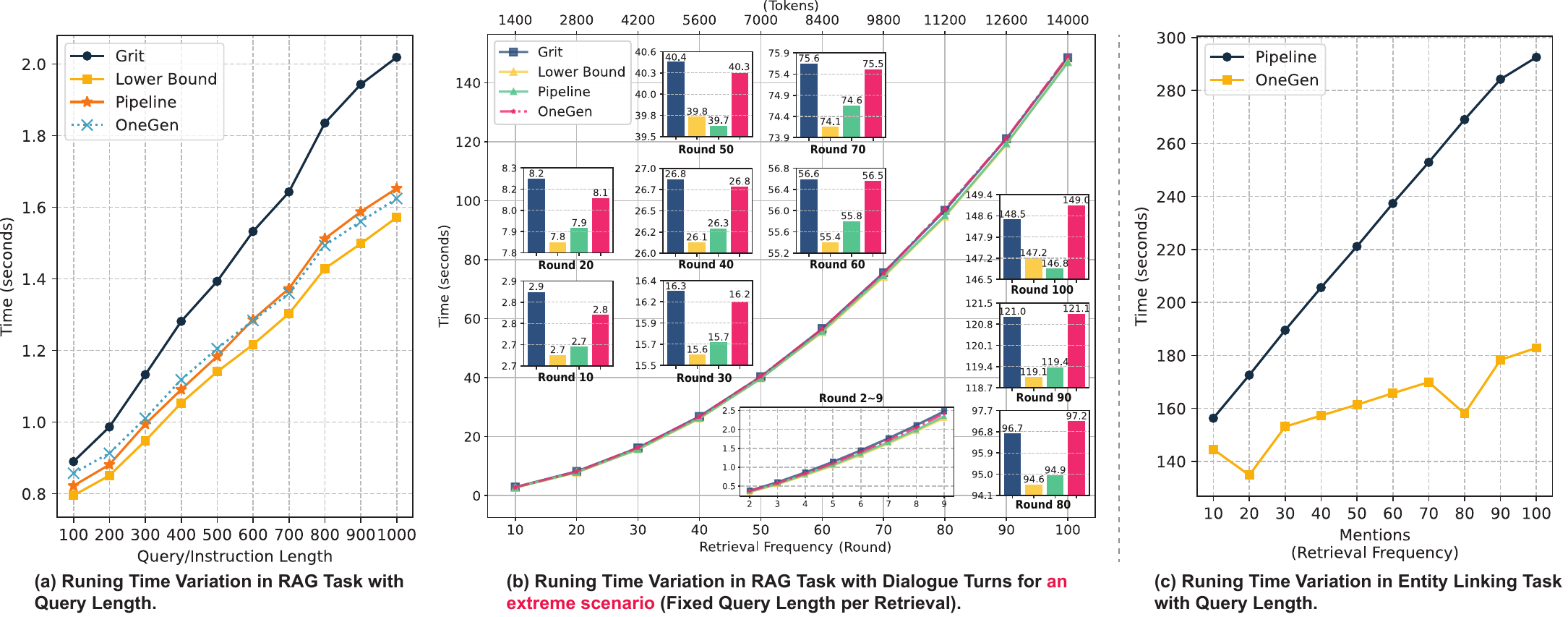} 
    }
    \caption{
    Efficiency analysis of \ours ~on RAG and Entity Linking tasks.
    All baselines maintain the same settings. 
    For RAG, the output is 10 tokens, with a document length of 30 tokens.
    Figure (a) illustrates the impact of query length on RAG efficiency across five dialogue rounds. 
    Figure (b) examines the influence of retrieval frequency and token length on RAG efficiency. 
    Figure (c) depicts how retrieval frequency affects efficiency in Entity Linking tasks.
    }
    \label{fig:exp:analysis:efficiency}
    \vspace{-2mm}
\end{figure*} 
To assess the efficiency of \ours~during inference, we evaluate the inference time for various scenarios and tasks, as illustrated in Figure~\ref{fig:exp:analysis:efficiency}.
For a fair comparison, all LLMs deployed are Mistral-v0.1-7B, operating with vLLM~\citep{sosp23_vllm} as the inference backend.
Following the optimum-benchmark\footnote{\url{https://github.com/huggingface/optimum-benchmark}}, all tokens are randomly generated and consistently numbered across baselines.
Notably, the ``Pipeline'' in Figure~\ref{fig:exp:analysis:efficiency}(a-b) employ Contriever as the retriever, while the Figure~\ref{fig:exp:analysis:efficiency}(c), following the EntGPT, the retriever in Pipeline is Mistral-v0.1-7B, converting ED task into a QA task.
The ``Lower Bound'' in Figure~\ref{fig:exp:analysis:efficiency}(a-b) denotes results obtained without retriever.
Detailed settings are provided in the Appendix~\ref{app:rg_single:experiments:efficiency} and \ref{app:gr:experiments:efficiency}.

$\textbf{R} \rightarrow \textbf{G Task.}$
Figure~\ref{fig:exp:analysis:efficiency}(a-b) depict inference latency comparisons, with the default instruction in each round serving as the query.
Figure~\ref{fig:exp:analysis:efficiency}(a) examines the influence of query length on inference speed in five rounds of $R\rightarrow G$.
Figure~\ref{fig:exp:analysis:efficiency}(b) assesses how the retrieval frequency and context length affect inference speed with a fixed query length of 100 for each round.
Key observations include:
\textit{1) Figure~\ref{fig:exp:analysis:efficiency}(a) shows that \ours's inference process is efficient, with a notably greater increase in speed as query length extends, compared to Grit.}
The efficiency stems from \ours's use of extra retrieval tokens, maintaining overall time close to that of smaller models (e.g. Contriever) used as retrievers in the Pipeline.
The improvement in inference speed with increasing query length varies from 4\% to 20\%, mainly due to \ours's elimination of a second query forward pass compared with the alternatives (pipeline \& GRIT).
\textit{2) Figure~\ref{fig:exp:analysis:efficiency}(b) reveals that \ours~maintains stable inference times, under extreme scenarios,} defined as one retrieval per dialogue round for 100 rounds with each round using 140 tokens and no semantic relevancy between rounds.
Even when dialogue rounds reach 80, the context length consequently extends to 11K tokens, the increase in inference latency is minimal, ranging from 0.1\% to 0.5\%, showcasing the \ours's stability.
In practical applications, other methods involve query rewriting on context for retrieval, leading to substantial overhead.
\ours, which can operate without an explicit query, enhancing efficiency significantly.

$\textbf{G} \rightarrow \textbf{R Task.}$
Figure~\ref{fig:exp:analysis:efficiency}(c) demonstrates the effect of mention count on inference speed for EL tasks in sentences of 1K tokens, where each mention equates to a retrieval instance.
Our findings include:
\textit{(1) \ours~is resource efficient,} deploying only one model compared to the dual-model setup in traditional pipelines.
\textit{(2) \ours~achieves enhanced inference efficiency, particularly as retrieval frequency increase.}
In scenarios ranging from 10 to 100 retrievals, \ours~reduces inference time by 8\% to 41\%.
This efficiency stems from bypassing the need to construct a query for each mention, unlike EntGPT.
Furthermore, as the inference operates under the Next Token Prediction paradigm, it benefits from advanced large model serving techniques such as vLLM.

\subsubsection{Impacts on Generation and Retrieval}
\label{sec:exp:impacts}
Here we examine whether situating retrieval and generation within the same context impacts the generative capacities of LLMs and assess the effectiveness of the retrieval.
Given that \S~\ref{sec:exp:main} evaluates end-to-end performance, where \ours~differs from other baselines in both the generation and retrieval modules, our evaluation principle here is to \textit{fix the retriever to assess generative capabilities} and \textit{fix the generator to evaluate retrieval capabilities.}
More details are shown in Appendix~\ref{app:rg_single:experiments:impact}

\begin{figure}[t]
\begin{minipage}{0.48\textwidth}
    \centering
    \small
    \scalebox{0.8}{
    \begin{tabular}{ccccc}
    \toprule
        \textbf{Method} & ReFinED & \begin{tabular}[c]{@{}c@{}}ChatEL \\ (GPT-4)\end{tabular} & \begin{tabular}[c]{@{}c@{}}EntGPT-I \\ (GPT-3.5)\end{tabular} & \begin{tabular}[c]{@{}c@{}}OneGen \\ (Llama2$_{\texttt{7B}}$)\end{tabular} \\ \midrule
    \textbf{AVG.} & 77.6    & 80.4           & {\underline{84.3}}         & \textbf{86.5}      \\ \bottomrule
    \end{tabular}
    }
    \vspace{-1mm}
    \captionof{table}{
    Impact of \ours ~on retrieval capabilities assessed through the ED task, presenting average F1 scores across nine datasets.
    Details are in Table~\ref{table:app:exp:gr:ed} in Appendix.
    }
    \label{table:exp:impact:gr:ed}
\end{minipage}\hfill
\begin{minipage}{0.48\textwidth} 
    \small
    \centering
    \vspace{-2mm}
    \scalebox{0.83}{
    \begin{tabular}{lccc}
    \toprule
    \textbf{Method} & ReFinED & \begin{tabular}[c]{@{}c@{}}Llama2$_{\texttt{7B}}$ \\ (SFT)\end{tabular} & \begin{tabular}[c]{@{}c@{}}Llama2$_{\texttt{7B}}$ \\ (\ours)\end{tabular} \\ \midrule
    \textbf{AVG.}   & 72.7    & 71.1   & 71.5   \\ \bottomrule
    \end{tabular}
    }
    \vspace{-1mm}
    \captionof{table}{Impact of \ours ~on generation capabilities assessed through the Mention Detection task, presenting average F1 scores across seven datasets.}
    \label{table:exp:impact:gr:md}
\end{minipage}
\vspace{-2mm}
\end{figure}

$\textbf{G} \rightarrow \textbf{R Task.}$
\textit{Same Retriever but different Generator:}
Evaluation is performed through the MD task.
Additionally, we train the LLM using the same data and hyperparameters with SFT (Supervised Fine-Tuning).
Table~\ref{table:exp:impact:gr:md} reports the average performance across seven datasets.
Comparing the last two columns of Table~\ref{table:exp:impact:gr:md}, we find that \textbf{\ours~does not impair the LLM's generative capabilities.}
\textit{Same Generator but different Retriever:}
We employ the Entity Disambiguation (ED) task to evaluate retrieval performance.
Table~\ref{table:exp:impact:gr:ed} summarizes the average results over nine datasets, revealing that \textbf{\ours~significantly enhances the retrieval capacity of LLMs.}

\subsubsection{Ablation Study}
The more ablation studies, such as $\lambda_r$,  implicit query, are shown in App.~\ref{app:rg_single:expriments:ablation}, ~\ref{app:rg_multi:experiments:ablation}, and ~\ref{app:gr:experiments:ablation}.

\textbf{Loss Function.}
We examine the impact of $\mathcal{L}_r$, comparing BPR and InfoNCE, on EL, MD, and ED task.
These tasks are assessed using seven, seven, and nine datasets respectively.
Our results, presented in Table~\ref{table:exp:ablation:gr:loss}, indicate the BPR consistently surpasses InfoNCE in performance.
This may be due to the overly restrictive of InfoNCE, which potentially limits the LLM's generative capabilities.

\section{Conclusion and Future Work}
\vspace{-2mm}
In this paper, we utilize mathematical notation to formally unify generative tasks, retrieval task, and their composites such as RAG and EL.
For composite tasks, we integrate retrieval and generation within the same context.
Building upon this unified approach, we propose the \ours~training framework, which harmonizes and expands both generative and representative instruction tuning.
We conduct extensive experiments on two distinct types of composite tasks, RAG and EL, to validate the pluggability, effectiveness, and efficiency of \ours~in training and inference. 
Furthermore, our results confirm that integrating generation and retrieval within the same context does not negatively impact the generative capabilities of LLMs, while also providing significant enhancements in retrieval capabilities.
Future research directions include:
1) Extending \ours~to the multimodal domain for tasks such as multimodal RAG and multimodal EL.
2) Enhancing \ours's training with diverse datasets to improve LLMs' capability for complex retrieval and generation tasks.
\newpage

\section*{Limitations}
Despite our comprehensive efforts, the study presents several limitations:

1) It remains unknown whether parameter-efficient fine-tuning methods such as LoRA~\citep{iclr22_LoRA} and QLoRA~\citep{nips23_QLoRA} could bring benefits for \ours~training. 
In this study, we utilized full-parameter fine-tuning. 
\ours~could potentially benefit from parameter-efficient fine-tuning, as recent work~\citep{arxiv24_e5_mistral} has used LoRA to equip LLMs with retrieval capabilities.

2) The absence of performance evaluations in more diverse and extensive data scenarios. 
Although we achieved gains with limited data, a more diverse set of tasks and data, such as jointly training with Entity Linking, RAG, retrieval data, and generation data, might produce a model with enhanced capabilities.

3) The efficacy of OneGen within Mixture of Experts (MoE) models has not been tested. 
It is possible that MoE architectures could significantly influence the routing of retrieval and generation tasks, potentially enhancing inference efficiency if integrated effectively with \ours.

4) The underlying mechanisms by which LLMs trained using \ours~achieve simultaneous retrieval and generation in a single forward pass, without mutual interference, remain unclear.

\section*{Acknowledgments}
We would like to express our sincere gratitude to the anonymous reviewers for their thoughtful and constructive feedback. 
We would like to thank Niels from HuggingFace for his valuable suggestions to improve the code.
This work was supported by the National Natural Science Foundation of China (No. 62206246, No. NSFCU23B2055, No. NSFCU19B2027), the Fundamental Research Funds for the Central Universities (226-2023-00138), Zhejiang Provincial Natural Science Foundation of China (No. LGG22F030011), Yongjiang Talent Introduction Programme (2021A-156-G), and Information Technology Center and State Key Lab of CAD\&CG, Zhejiang University. 
This work was supported by Ant Group and Zhejiang University - Ant Group Joint Laboratory of Knowledge Graph.

\bibliography{iclr2024_conference}
\bibliographystyle{iclr2024_conference}

\clearpage
\appendix

\section*{Appendix}

\startcontents[sections]
\printcontents[sections]{l}{1}{\setcounter{tocdepth}{3}}
\newpage

\section{Related Works}
\label{app:related_work}
\begin{figure}[!hb] 
    \centering
    \scalebox{0.9}{
    \includegraphics[width=1\linewidth]{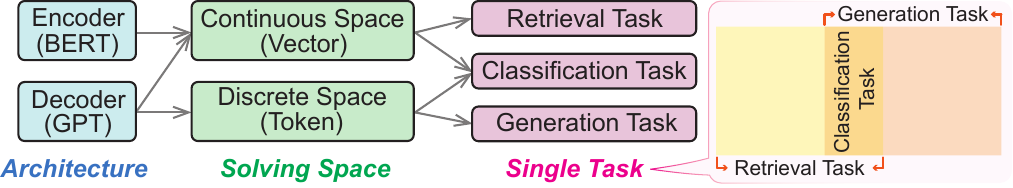} 
    }
    \caption{The relation between the model architecture, solving space, and tasks. }
    \label{fig:app:related_works:overview}
\end{figure} 
Natural Language Processing (NLP) tasks can be solved in two types of \textbf{spaces}: \textit{continuous} and \textit{discrete}. 
Encoder architectures (e.g., BERT~\citep{naacl19_BERT}) operate in the \textit{continuous space}, where the basic units of processing are continuous \textit{vectors}. 
Decoder architectures (e.g., GPT~\citep{nips20_gpt}) and Encoder-Decoder architectures (e.g., T5~\citep{jmlr20_T5}) typically operate in the discrete space, processing discrete \textit{tokens} as their fundamental units.

NLP \textbf{tasks} consist of two major categories: \textit{Natural Language Understanding} (NLU) and \textit{Natural Language Generation} (NLG). 
NLU tasks often involve classification (e.g., sentiment classification), while NLG tasks typically involve generation (e.g., novel generation). 
Specifically, classification tasks with a large number of classes are referred to as retrieval tasks. 
For clarity, we define tasks with fewer classes as classification tasks, and those with many classes as retrieval tasks. 
Generally, classification tasks can be reframed as generation tasks.

Figure~\ref{fig:app:related_works:overview} illustrates the relationship between NLP tasks and the spaces in which they are solved. 
As NLP has advanced, various composite tasks have emerged, such as Retrieval Augmented Generation (RAG) tasks and Entity Linking (EL), which often require coordination between retrieval and generation. 
These composite tasks can be divided into two categories based on whether retrieval serves generation: $R\rightarrow G$ tasks and $G\rightarrow R$ tasks. 
RAG is a representative $R\rightarrow G$ task, while EL is a $G\rightarrow R$ task, where the retrieved content does not directly serve generation.

For composite tasks, the generation task is always handled by a LLM.
Therefore, we first introduce work related to LLM-based retrieval, and then we discuss related work on composite tasks.

\subsection{LLM-based Retrieval}
\label{app:related_works:retrieval}
Numerous works have utilized LLMs to perform retrieval tasks. 
These can be categorized into continuous space retrieval and discrete space retrieval, depending on the unit used for retrieval.
As shown in Figure~\ref{fig:app:related_works:overview}, \textit{discrete space retrieval cannot directly search from a vast candidate pool.} 
Therefore, it typically relies on recall methods to narrow down the search space. 
In contrast, \textit{vector space retrieval can directly identify the target set from a large candidate pool.}
\paragraph{LLM-based Retrieval in Continuous Space.}
Continuous space is well-suited for retrieval tasks. 
The core of retrieval lies in how documents or queries are encoded. 
We classify the methods based on whether training is required and the type of attention mechanism used:
\begin{itemize}
\item{
\textbf{Prompt-based methods using Causal Attention.}
PromptEOL~\citep{arxiv23_PromptEOL} encodes documents by carefully crafting prompts and using logits from specific positions. 
EchoEmbedding~\citep{arxiv24_EchoEmb} achieves document encoding by repeating the document and extracting logits from the final token or summing logits from corresponding positions. 
These methods demonstrate that current LLMs without training possess a certain degree of retrieval capability.
}
\item{
\textbf{Trained methods using Causal Attention.}
E5-Mistral~\citep{arxiv24_e5_mistral} encodes documents by training on synthetic data, using the logits of the document’s final token. 
EchoEmbedding~\citep{arxiv24_EchoEmb} also provides a method for supervised training on logits of repeated documents.
}
\item{
\textbf{Trained methods using Bidirectional Attention.}
GritLM~\citep{arxiv2024_GritLM} and LLM2VEC~\citep{arxiv24_llm2vec} replace the causal attention mechanism in LLMs with bidirectional attention and use Mean Pooling to obtain document encodings. 
GritLM is directly trained on supervised data, while LLM2VEC employs masked next-token prediction training with unsupervised data.
}
\end{itemize}

\paragraph{LLM-based Retrieval in Discrete Space.}
Discrete space retrieval is computationally intensive. 
It is commonly achieved through constrained decoding processes or structured as QA tasks:
\begin{itemize}
\item{
\textbf{Constrained decoding.}
Approaches like GENRE~\citep{iclr21_GENRE} and RICH~\citep{arxiv_deepmind_riches} first generate a corresponding query with the LLM, which is then used to recall candidates from a pre-built index (such as a trie tree or FM-Index). 
The LLM then scores these candidates, guiding subsequent outputs to align with a specific document from the recalled set.
Beam search is a main technique to implement the above process, where beam size is large and is often set to 10.
}
\item{
\textbf{QA-based.}
In these methods, the recalled candidate is typically concatenated directly into the text, instructing the LLM to output the index of the most relevant document or assign scores to each document, as seen in methods like RankRAG~\citep{arxiv24_rankrag}.
}
\end{itemize}

\subsection{Composite Task}
\begin{figure*}[] 
    \centering
    \scalebox{1}{
    \includegraphics[width=1\linewidth]{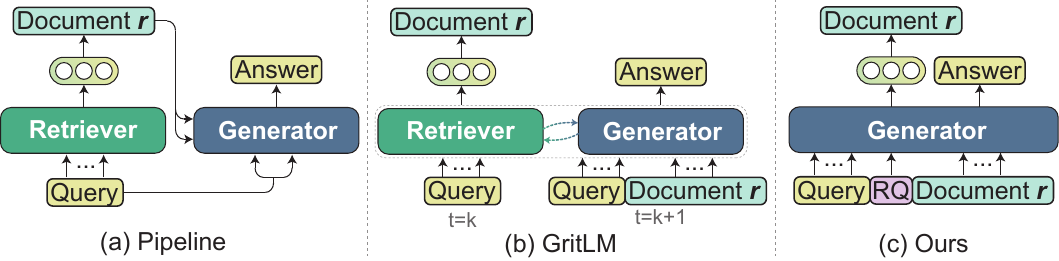} 
    }
    \caption{
    Comparison of three methods for completing RAG task.
    }
    \label{fig:app:comparison:rag}
\end{figure*}

In this section, we define methods that require deploying two models during inference as \textit{Pipeline} methods, regardless of whether these models are trained jointly or separately.
Conversely, methods that rely on a single model are termed \textit{Single Model} methods.
It is important to note that if a method does not require the deployment of additional neural networks at the retrieval level and instead utilizes techniques such as BM25, TRIE trees, or FM-Index for retrieval, we classify these methods as belonging to the \textit{Single Model} category.
\paragraph{Pipeline.}
The basic workflow of pipeline methods in Retrieval-Augmented Generation (RAG) is illustrated in Figure~\ref{fig:app:comparison:rag} (a).
Typically, these methods involve deploying an extra model for retrieval followed by a large language model (LLM) for generation. 
This approach is prevalent in many works, including Self-RAG~\citep{iclr24_selfrag}, ChatQA~\citep{arxiv24_chatqa}, RankRAG~\citep{arxiv24_rankrag}, and RQ-RAG~\citep{arxiv24_RQ-RAG}. 
Since pipeline methods require the use of two separate models, query generation is essential. 
The LLM-generated query must be input into the retrieval model, necessitating two forward computations of the query. 
Specifically, Self-RAG introduces various scoring tokens to provide feedback on both the retrieved content and the generated output, optimizing retrieval augmentation. 
RQ-RAG enhances multi-step reasoning in question answering by decomposing the user’s original query using Chain-of-Thought (CoT~\citep{nips22_cot}) reasoning,. 
RankRAG improves retrieval performance by incorporating document ranking task during training, while ChatQA enhances the model's ability to understand contextual information, leading to better answer generation, as shown in Table~\ref{table:comparision}.

\paragraph{Single Model.}
Single model methods significantly reduce memory overhead but lack a unified approach. 
As shown in Figure~\ref{fig:app:related_works:overview}, we categorize existing single model methods into three types:
\begin{itemize}
\item{
\textbf{Encoder$\rightarrow$Continuous Space.}
Due to its encoder architecture, this approach cannot directly perform generation tasks. 
Entity Linking (EL) task involves Mention Detection (MD) and Entity Disambiguation (ED). 
ReFinED~\citep{naacl22_refined}, although using two models during training—one for encoding documents and another for encoding queries and sequence labeling—aligns these encoders in representational space, so only the query encoder is required during inference. 
Thus the documents are encoded offline and cached. 
ReFinED handles Mention Detection as a sequence labeling task, effectively solving it as a classification problem. 
For Entity Disambiguation, ReFinED uses mean pooling on mentions and retrieves from a cached document embeddings. 
However, the encoder architecture's limitations restrict its application to other tasks.
}
\item{
\textbf{Decoder$\rightarrow$Discrete Space.}
Given the Decoder architecture and the necessity of retrieval in a discrete space, the recall process becomes indispensable.
Methods like GENRE~\citep{iclr21_GENRE} and RICH~\citep{arxiv_deepmind_riches} address this by constructing a TRIE tree and an FM-Index on the documents, respectively.
These approaches perform retrieval concurrently with generation, as explained in Appendix~\ref{app:related_works:retrieval}. 
However, the primary limitation for these approaches is time costing, often involving a beam search with a beam size of 10.
Moreover, constructing training data for these approaches are stringent, requiring precise annotations for each retrieval point, unlike our method (see Appendix~\ref{app:feature}).
}
\item{
\textbf{Encoder$\rightarrow$Continuous Space \& Decoder$\rightarrow$Discrete Space.}
GritLM~\citep{arxiv2024_GritLM} uses bidirectional attention to encode queries and documents during retrieval and causal attention during generation. 
This switching leads to two outcomes: 
1) The query requires two forward computations since encoding and generation use different attention mechanisms, as shown in Figure~\ref{fig:app:comparison:rag}(b). 
2) The LLM's key-value cache cannot be effectively utilized.
}
\end{itemize}

\section{Relation to prior work on LLM instruction tuning} 
\label{app:priorwork}
\begin{table*}[!htp]
\centering
\small
\scalebox{0.9}{
\begin{tabular}{lcccc}
\toprule
\multirow{2}{*}{\textbf{Method}}  & \multirow{2}{*}{\textbf{Loss}} & \multicolumn{3}{c}{\textbf{Supported Data} ($role(x_i) \in \{?\}$)} \\ \cmidrule{3-5}
     &                       & $\{\texttt{CTX},\texttt{GEN}\}$ & $\{\texttt{CTX},\texttt{RET}\}$  & $\{\texttt{CTX},\texttt{RET}, \texttt{GEN}\}$ \\ \midrule
GIT (SFT)  & $\mathcal{L}=\mathcal{L}_g$ & \checkmark & \ding{55} & \ding{55}\\
RIT  & $\mathcal{L}=\mathcal{L}_r$ & \ding{55}   & \checkmark  & \ding{55} \\
GRIT & $\mathcal{L}=\lambda_g\mathcal{L}_g+\lambda_r\mathcal{L}_r$ & \checkmark  & \checkmark & \ding{55} \\ \midrule
\ours  & $\mathcal{L}=\lambda_g\mathcal{L}_g+\lambda_r\mathcal{L}_r$ & \checkmark  & \checkmark & \checkmark \\ \bottomrule
\end{tabular}
}
\caption{Comparison of four Instruction Tuning}
\label{table:comparision}
\end{table*}
Recent work on representative instruction fine tuning (RIT) \citep{sigir24_ft_llama, arxiv24_e5_mistral} demonstrated great potential for autoregressive LLMs in constructing high-quality embedding. 
They often use the logits (e.g. hidden states in the last layer) of $EOS$ token appended at the end of the sentence to represent the input sequence. 
However, RIT models are only used for retrieval tasks due to the degenerated performance on generation after fine-tuning \citep{arxiv2024_GritLM}. Grit~\citep{arxiv2024_GritLM} uses different instruct prompts to control the switch between generation and embedding within a single LLM fine-tuned for both generation and representation instruction tuning and showed unifying the training task could improve the generation performance. 
The need for different prompts, however, makes it less efficient in applications requiring interleaved generation and retrieving. 
\textbf{From the perspective of training methods, \ours~is an extension of GIT and RIT, and it can degenerate to GIT and RIT.}
\section{Broader Application}
\label{app:application}
Tasks that require both \textit{generation} and \textit{retrieval} (or \textit{classification}) can be effectively handled using \ours.
Our approach enables the LLM to perform retrieval and classification during the generation process without any modification of architecture.
Beyond the experimental tasks detailed in the main text, such as retrieval-augmented multi-hop reasoning and entity linking, here are two additional potential applications:
\begin{itemize}
    \item{
    \textbf{Text to Linked Knowledge Graph Triples.}
    Traditional knowledge graph construction typically involves several steps, including Named Entity Recognition (NER), Relation Extraction (RE), and Entity Linking (EL).
    With the advent of LLMs, it's possible to directly generate triples without linking from the sentence or document.
    Using \ours, we can enhance this process by adding a special token (e.g., \texttt{[RQ]}) after each entity in the training data, allowing the LLM to generate linked knowledge graph triples in a single forward pass.
    }
    \item{
    \textbf{Controlled Generation Scenarios.}
    An example is KTS~\citep{arxiv24_KTS}, a method designed to prevent harmful outputs from LLM.
    Specifically, it classifies user inputs in advance.
    If the input is harmful, a pre-constructed steering vector~\citep{arxiv23_actadd} is inserted to ensure a safe response.
    Otherwise, the steering vector is not inserted.
    In implementation, by simply appending a special token (e.g., \texttt{[RQ]}) to each prompt, we can meet these requirements without relying on external classifiers or redundant encoding processes.
    }
\end{itemize}

\section{Why does \ours~work?}
\label{app:why}
From a training perspective, \ours~training approach is similar to that of multi-turn dialogue, where not all tokens contribute to the next token prediction loss.
From an inference perspective, the multi-turn dialogue of LLM extends the new conversation by directly appending the user's next question after each dialogue round, whereas \ours~addresses this issue by appending a \texttt{<CON>} token. 
If we define the \texttt{[RQ]} token as the end of each dialogue round, the primary distinction between \ours training is that, in multi-turn dialogues, the last token of each round (such as \texttt{<|im\_end|>} in Qwen2, \texttt{</s>} in Llama2, \texttt{<|eot\_id|>} in Llama3) does not contribute to any loss during training and is thus redundant.
In contrast, we leverage this token to encode the current intent.
Therefore, since the multi-turn dialogue training method is effective, our approach is also effective.

So why is it possible to encode a query with just one token in the process of generation?
First, LLMs inherently possess encoding capabilities, which can be enhanced with simple training to improve the LLM's retrieval abilities.
Recent prompt-based methods, such as EchoEmbedding~\citep{arxiv24_EchoEmb} and PromptEOL~\citep{arxiv23_PromptEOL}, have demonstrated that LLMs can achieve good retrieval results even without additional training.
We believe that \textit{an excellent painter (capability of generation) must have a great sense of aesthetics (capability of understanding), but the reverse is not necessarily true.}
Second, many studies~\citep{emnlp23_llmlingua} have compressed prompts into a single token.
Therefore, we can view special tokens, such as \texttt{[RQ]} and \texttt{[RD]}, as compressed representations of the prompts ``\texttt{represent the current query}'' and ``\texttt{represent the current document}''.

\section{\ours's Features}
\label{app:feature}

\textbf{Support for Diverse Training Data.}
\ours~can handle various mixed data types, as shown in Table~\ref{table:comparision}.

\textbf{Pluggability.}
It is adaptable to different LLM architectures and sizes and can be integrated with existing methods such as Self-RAG~\citep{iclr24_selfrag} and RQ-RAG~\citep{arxiv24_RQ-RAG}.

\textbf{Training Efficiency.}
 The use of the BPR~\citep{uai09_BPR} loss function for optimizing $\mathcal{L}_r$ allows for Gradient Accumulation, which enables larger batch sizes for contrastive learning.
 Additionally, \ours~achieves competitive performance with less training data.

\textbf{Inference Efficiency.}
\ours~avoids redundant computations by requiring only a single forward pass for queries, thereby reducing additional processing. 
Its unchanged model structure facilitates the use of existing inference acceleration techniques, such as vLLM~\citep{sosp23_vllm}.

\textbf{No Need for Query Rewriting or Query Generation.}
In the Multi-Hop QA setting, we demonstrate that the model can achieve comparable results to the baseline without generating or rewriting queries. 
Refer to Appendix~\ref{app:rg_multi:experiments:ablation} for details.

\textbf{Flexible Annotation Requirements.}
    \ours's retrieval operates in continuous space, meaning that annotations for retrieval components in training data do not need to be highly precise.
    For example, in the EL task, given an expected output ``\texttt{<LOC>Steve Jobs</LOC> [RQ] <CON> founded <LOC>Apple Inc</LOC> [RQ] <CON>}'', we may need to find the entities corresponding to``Steve Jobs'' and ``Apple Inc'' in KB. 
    However, for \ours, it is sufficient to annotate only ``Steve Jobs'' or ``Apple Inc'', or even omit annotations altogether.
    The LLM can use annotations from other training data to optimize the \texttt{[RQ]} token.

\clearpage
\section{$\textbf{R}\rightarrow \textbf{G}$ Task: RAG for Single-hop QA}
\label{app:rg_single}
This section presents the details of the $\textbf{R}\rightarrow \textbf{G}$ task, including the detailed construction of the data, training details, inference details, and additional experimental results.
Figure~\ref{fig:app:comparison:rag} shows a comparison of three methods for completing the RAG task.
\subsection{Introduction of Self-RAG}
\label{app:rg_single:self_rag}
Self-RAG~\citep{iclr24_selfrag} is a method designed for adaptive retrieval and self-assessment.
Specifically, when a query input to the LLM, it begins generating responses in an autoregressive manner.
When the LLM outputs the \texttt{[Retrieve]} token, it halts generation and invokes a retriever (e.g., Contriever~\citep{tmlr22_contriever}) to retrieve relevant documents.
These documents are then appended to the current context using the template ``\texttt{\{history\}<paragraph>{relevant document}</paragraph>}''.
The LLM is then required to output either \texttt{[Relevant]} or \texttt{[Irrelevant]} to determine if the retrieved documents are relevant to the query.
If the documents are deemed irrelevant, the LLM outputs the query again and performs another retrieval.
If relevant, the LLM continues to generate the answer.
Self-RAG employs tokens such as \texttt{[Fully Supported]}, \texttt{[Partially Supported]}, and \texttt{[No Support]} to evaluate whether the generated answer aligns with the retrieved documents, and uses rating tokens like \texttt{[Utility:5]}, \texttt{[Utility:4]}, \texttt{[Utility:3]}, \texttt{[Utility:2]}, and \texttt{[Utility:1]} to assess the relevance of the answer to the original question.
For more details, we refer readers to the original paper.

\subsection{Training Details}
\label{app:rg_single:training}
\begin{figure*}[] 
    \centering
    \scalebox{1}{
    \includegraphics[width=1\linewidth]{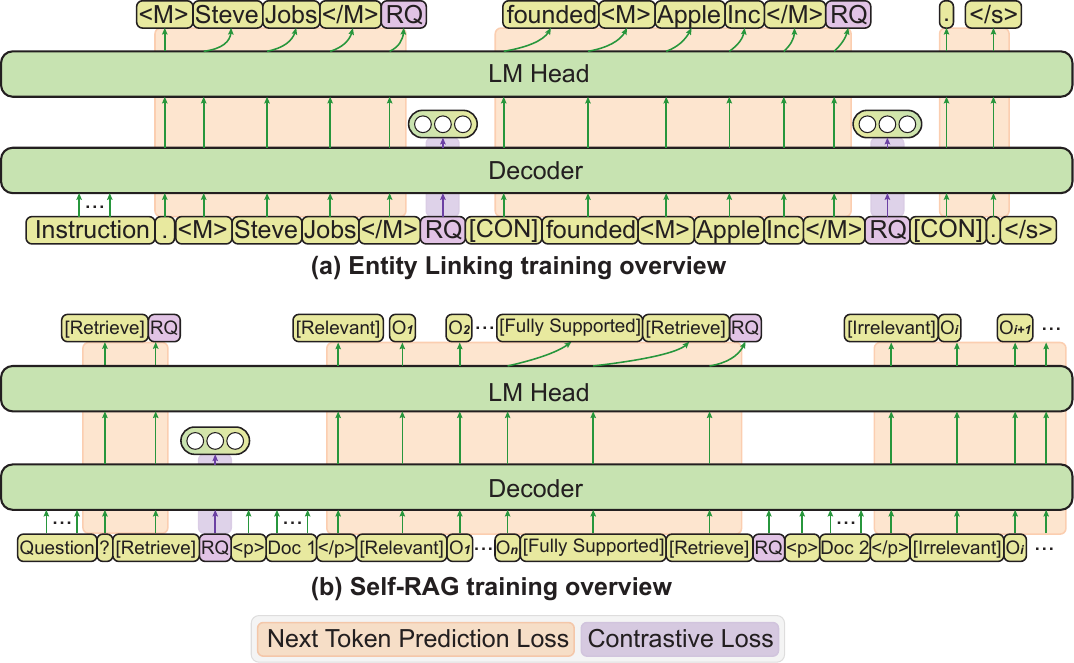} 
    }
    \caption{
    The detailed training process of Entity Linking and Self-RAG.
    }
    \label{fig:app:training}
\end{figure*}

A comprehensive example of training Self-RAG is illustrated in Figure~\ref{fig:app:training} (b).

\begin{figure*}[h] 
    \centering
    \scalebox{1}{
    \includegraphics[width=1\linewidth]{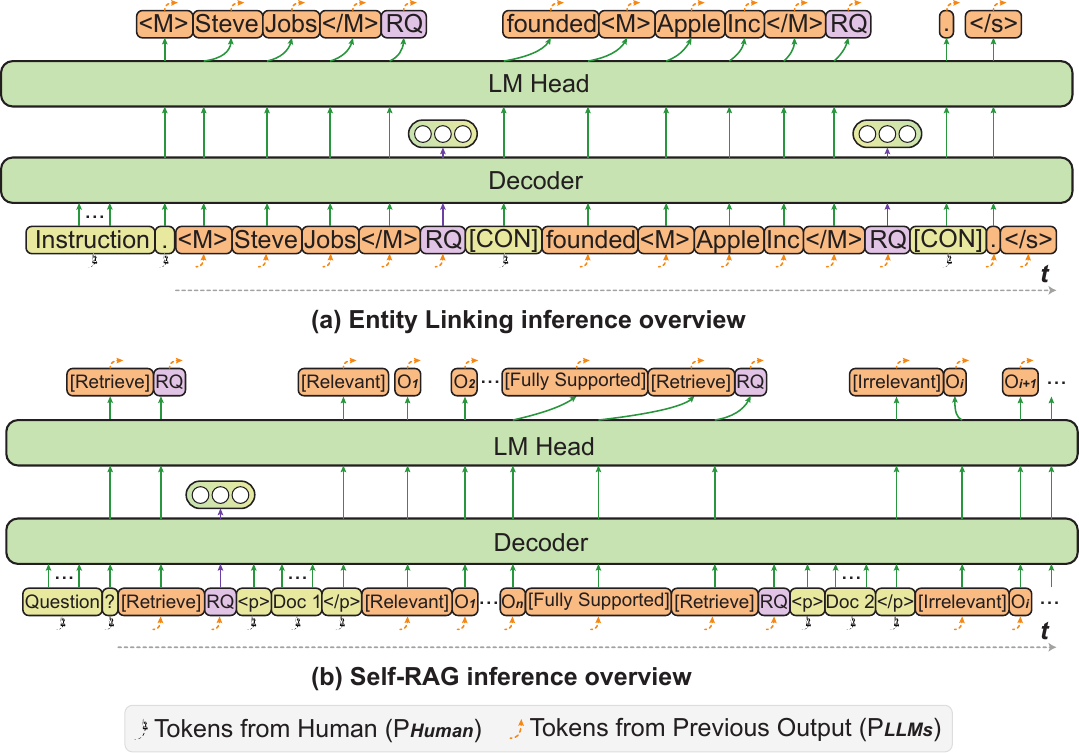} 
    }
    \caption{
    The detailed inference process of Entity Linking and Self-RAG.
    }
    \label{fig:app:inference}
\end{figure*} 
\textbf{Data Reconstruction.}
Here, we outline the modifications we implemented in the training dataset for Self-RAG~\citep{iclr24_selfrag}. 
The following diagram presents a typical Self-RAG training example, featuring special tokens highlighted in blue and red.
During training, all tokens except for the black text are involved in computing the generative loss.
\begin{tcolorbox}
\textbf{Input:}

Q: Where do you work?

A: Apple Inc., the iPhone is one of their products.

Q: Who is the CEO?

\textbf{Output:}

\textcolor{blue}{\texttt{[Retrieval]}}
\textcolor{red}{\texttt{<p>}}
\textcolor{black}{Timothy Cook (born November 1, 1960) is an American business executive who is the current chief executive officer of Apple Inc.}
\textcolor{red}{\texttt{</p>}}
\textcolor{blue}{\texttt{[Relevant]}}
\textcolor{brown}{Tim Cook}
\textcolor{blue}{\texttt{[Fully Supported]}}
\textcolor{red}{\texttt{[Utility:5]}}.
\end{tcolorbox}
Take this for example, we introduce a \texttt{[RQ]} token immediately following the \textbf{[Retrieval]} token (illustrated as the purple token in the following diagram), enabling the LLM to extract the semantic information of the query at this point, with the rest of the setup remaining unchanged. 
In training, on top of the original setting, we compute a representative loss specifically for the purple token, e.g.,$role(\texttt{[RQ]})=\texttt{RET}$. 
It should be noted that a \texttt{[RQ]} token is appended after every \texttt{[Retrieval]} token. However, not all are included in the training process.
Specifically, if a sequence formatted as \texttt{[Retrieval][RQ]<p>$\cdots$</p>[Irrelevant]} occurs, indicating that the document encapsulated within \texttt{<p>$\cdots$</p>} is followed by a \texttt{[Irrelevant]} token, then the \texttt{[RQ]} token in this context does not contribute to the calculation of loss.

\begin{tcolorbox}
\textbf{Input:}

Q: Where do you work?

A: Apple Inc., the iPhone is one of their products.

Q: Who is the CEO?

\textbf{Output:}

\textcolor{blue}{\texttt{[Retrieval]}}
\textcolor{violet}{\textbf{\texttt{[RQ]}}}
\textcolor{red}{\texttt{<p>}}
\textcolor{black}{Timothy Cook (born November 1, 1960) is an American business executive who is the current chief executive officer of Apple Inc.}
\textcolor{red}{\texttt{</p>}}
\textcolor{blue}{\texttt{[Relevant]}}
\textcolor{brown}{Tim Cook}
\textcolor{blue}{\texttt{[Fully Supported]}}
\textcolor{red}{\texttt{[Utility:5]}}.
\end{tcolorbox}
The calculation of the representative loss requires positive and negative samples.
Hence, for the positive samples, we use the content enclosed within \texttt{<p>...</p>} tags. 
For the negative samples, we utilize the Mistral-E5~\citep{arxiv24_e5_mistral} to embed the document corpus~\footnote{\url{https://dl.fbaipublicfiles.com/dpr/wikipedia_split/psgs_w100.tsv.gz}} and select documents ranked from 5985 to 6000 as negative samples. 
The numbers 5985 and 6000 are hyperparameters for constructing negative samples. Since the Self-RAG training data does not include negative samples (as it uses the Contriever, which doesn't require training), and \ours~needs to train retrieval capabilities, we have to create negative samples. In the document pool provided by Self-RAG, documents are divided into different chunks, and multiple chunks or documents may correspond to the answer for a given query. This means that not only one document necessarily contains the correct answer. To avoid selecting false negatives, we decided to use chunks ranked between 5985 and 6000 as negative samples.
Examples of both positive and negative samples are provided below.
\begin{figure}[h]
\begin{tcolorbox}
\textbf{Positive Example:}

\textcolor{black}{Timothy Cook (born November 1, 1960) is an American business executive who is the current chief executive officer of Apple Inc.}\textcolor{violet}{\textbf{\texttt{[RD]}}}

\textbf{Negative Example:}

\textcolor{black}{After graduating from Auburn University, Cook spent twelve years in IBM's personal computer business, ultimately as director of North American fulfillment.}\textcolor{violet}{\textbf{\texttt{[RD]}}}
\end{tcolorbox}
\vspace{-3mm}
\caption{A case for data reconstruction of positive document and negative document about Self-RAG. Add \texttt{[RD]} token at the end of each document.}
\label{app:fig:rg:self_rag:doc:case}
\end{figure}
\vspace{-2mm}

\textbf{Implementation Details.}
The original dataset for Self-RAG~\citep{iclr24_selfrag} training comprises 150k instances, of which 60k are suitable for computing the representative loss. 
We perform comprehensive training on eight A800 machines utilizing the DeepSpeed ZeRO-3~\citep{sc20_deepspeed} strategy for memory efficiency.
We set the gradient accumulation to 4, and the batch size per GPU is 3, resulting in a final global batch size of $8\times 3\times 4=96$.
Training is conducted over 3 epochs with a learning rate set at 2e-5 and a 3\% warm-up period.
Both $\lambda_g$ and $\lambda_r$ are set at 1.
Data is sampled randomly.
Within a batch, the loss $\mathcal{L}_r$ is computed if the \texttt{[RQ]} token is present.
Otherwise, it is omitted.
For each \texttt{[RQ]} token, we sample one positive and two negative documents, with negative documents shared across the batch.

\subsection{Inference Details}
\label{app:rg_single:inference}
Algorithm~\ref{algo:self-rag} presents the pseudocode for Self-RAG inference.
Figure~\ref{fig:app:inference}(b) provides a schematic representation of the Self-RAG inference process.

\renewcommand{\thealgorithm}{1}
\label{algo:self-rag}
\begin{algorithm*}
\caption{RAG Inference}
\begin{algorithmic}[1]
\Require 
\Statex LLM trained with OneGen, denoted as $\hat{f}(\cdot)$
\Statex LLM without the LM-Head, denoted as $f(\cdot)$
\Statex Pre-cached document vector library $Emb_{doc}$
\Statex Instruction $x$
\Statex Cosine similarity computation function \texttt{CosineSimilarity()}
\Statex Function to sort and return the corresponding documents \texttt{Top1Doc()}

\Ensure
\Statex Answer $History$

\Statex\hspace*{-\algorithmicindent}\hrulefill
\State $History \gets x$
\State $NextToken \gets \hat{f}(History)$
\While{$NextToken \notin \texttt{Terminator}$}
    \State $History \gets History \cup \{NextToken\}$
    \If{$role(NextToken) = \texttt{RET}$}\textcolor{blue}{{\Comment{Retrieval on demand}}}
        \State $scores \gets \texttt{CosineSimilarity}(f(History), Emb_{doc})$
        \State $\texttt{RelevantDoc}\gets\texttt{Top1Doc}(scores)$
        \State $History \gets History \cup \{\texttt{RelevantDoc}\}$
    \EndIf
    \State $NextToken \gets \hat{f}(History)$\textcolor{blue}{{\Comment{Generation}}}
\EndWhile
\State \Return $History$
\end{algorithmic}
\end{algorithm*}
\subsection{Evaluation Details}
\label{app:rg_single:evaluation}
\textbf{Baselines.}
Baselines can be categorized into three types.
The first, named \textit{LLMs with proprietary data}, involves directly questions into strong models like ChatGPT, without using any retrieval documents or specialized training.
The second named \textit{Baselines without retrieval} uses LLMs such as Llama2 and Alpaca in 7B-Chat and 13B-Chat configurations.
The third named \textit{Baselines with retrieval}. 
Except for Llama2-7B-FT~\citep{arXiv2023_LLaMA}, which is finetuned using Self-RAG training data without reflection token, responses are generated from the concatenation of questions and retrieved documents
All the baselines use Contriever-MSMARCO~\citep{tmlr22_contriever} as the retriever, except for the GritLM-7B, which serves as its own retriever.

\textbf{Evaluation Setup, Datasets, and Metric.}
We evaluated \ours~on four datasets: PubHealth~\citep{arxiv23_PubHealth}, ARC-Challenge~\citep{arxiv18_ARC}, PopQA~\citep{acl23_PopQA} and TriviaQA~\citep{acl17_TriviaQA}.
For the first two, we use accuracy as an evaluation metric.
For the others, we assess performance by checking if the model generations contain gold answers, rather than insisting on exact matches, as per the method used by ~\cite{acl23_PopQA}; ~\cite{iclr24_selfrag}; ~\cite{nips23_Toolformer}.
Moreover, each candidate document corresponding to a question is provided by its respective dataset.

\subsection{Experiments}
\label{app:rg_single:experiments}
\begin{table*}[!t]
\centering
\small
\scalebox{0.8}{
\begin{tabular}{cccclccccc}
\toprule
\multirow{3}{*}{\textbf{LLMs}} & \multicolumn{3}{c}{\textbf{Retriever}}           &  & \multicolumn{4}{c}{\textbf{Dataset}}                                   & \multirow{3}{*}{\textbf{AVG.}} \\ \cmidrule{2-4} \cmidrule{6-9}
                      & Name       & Dataset Name & Dataset Size      &  & PopQA         & TQA           & Pub           & ARC           &                      \\ \midrule
\multicolumn{10}{c}{\textit{LLMs with proprierary data}}                                                                                                           \\ \midrule
Llama2-$c_\texttt{13B}$           & -          & -         & -              &  & 20.0          & 59.3          & 49.4          & 38.4          & 41.8                 \\
Ret-Llama2-$c_\texttt{13B}$       & -          & -         & -              &  & 51.8          & 59.8          & 52.1          & 37.9          & 50.4                 \\
ChatGPT               & -          & -         & -              &  & 29.3          & 74.3          & 70.1          & 75.3          & 62.3                 \\
Ret-ChatGPT           & -          & -         & -              &  & 50.8          & 65.7          & 54.7          & 75.3          & 61.6                 \\ \midrule
\multicolumn{10}{c}{\textit{Baselines without retrieval}}                                                                                                          \\ \midrule
Llama2$_{\texttt{7B}}$~\citep{arXiv2023_LLaMA}            & -          & -         & -              &  & 14.7          & 30.5          & 34.2          & 21.8          & 25.3                 \\
Alpaca$_{\texttt{7B}}$~\citep{nips23_Alpaca}            & -          & -         & -              &  & 23.6          & 54.5          & 49.8          & 45.0          & 43.2                 \\
Llama2$_{\texttt{13B}}$~\citep{arXiv2023_LLaMA}           & -          & -         & -              &  & 14.7          & 38.5          & 29.4          & 29.4          & 28.0                 \\
Alpaca$_{\texttt{13B}}$~\citep{nips23_Alpaca}           & -          & -         & -              &  & 24.4          & 61.3          & 55.5          & 54.9          & 49.0                 \\ \midrule
\multicolumn{10}{c}{\textit{Baselines with retrieval}}                                                                                                             \\ \midrule
Toolformer~\citep{nips23_Toolformer}            & Contriever & MS MARCO  & $1\times 10^{6}$             &  & -             & 48.8          & -             & -             & -                 \\
Llama2$_{\texttt{7B}}$~\citep{arXiv2023_LLaMA}             & Contriever & MS MARCO  & $1\times 10^{6}$             &  & 38.2          & 42.5          & 30.0          & 48.0          & 39.7                 \\
Alpaca$_{\texttt{7B}}$~\citep{nips23_Alpaca}             & Contriever & MS MARCO  & $1\times 10^{6}$             &  & 46.7          & 64.1          & 40.2          & 48.0          & 49.8                 \\
SAIL$_{\texttt{7B}}$~\citep{arxiv23_SAIL}             & Contriever & MS MARCO  & $1\times 10^{6}$             &  & -          & -         & 69.2          & 48.4          & -                 \\
Llama2-FT$_{\texttt{7B}}$~\citep{arXiv2023_LLaMA}               & Contriever & MS MARCO  & $1\times 10^{6}$             &  & 48.7          & 57.3          & 64.3          & 65.8          & 59.0                 \\
Mistral$_{\texttt{7B}}$~\citep{arXiv2023_Mistral}             & Contriever     & MS MARCO      & $1\times 10^{6}$ &  & {23.2} & {49.3} & 52.0          & 39.0          & 40.9                 \\ 
GritLM$_{\texttt{7B}}$~\citep{arxiv2024_GritLM}             & GritLM$_{\texttt{7B}}$     & E5S(w/ TQA)       & $2\times 10^{6}$ &  & \textbf{58.0} & \textbf{66.5} & 49.7          & 24.5          & 49.7                 \\ \midrule
Self-RAG$_{\texttt{7B}}$~\citep{iclr24_selfrag}            & Contriever & MS MARCO  & $1\times 10^{6}$             &  & {\underline{52.5}}    & 65.0          & {\underline{72.2}}    & {\underline{67.3}}    & {\underline{64.3}}           \\
Self-RAG$_{\texttt{7B}}$~(+\ours)      & \textit{Self}       & \textit{Sampled}    & \textbf{$\bf 6\times 10^{4}$}            &  & {\underline{52.5}}    & {\underline{65.7}}    & \textbf{75.1} & \textbf{70.1} & \textbf{65.8}        \\ \bottomrule
\end{tabular}
}
\vspace{-1mm}
\caption{
Performance comparison across different datasets.
Best and second-best results within `Baselines with retrieval' are indicated in bold and underlined, respectively. 
}
\label{table:exp_rag_all}
\vspace{-1.5mm}

\end{table*}
\subsubsection{Efficiency Settings}
\label{app:rg_single:experiments:efficiency}
We conduct tests on a single 40GB A100 card, with an \texttt{Intel(R) Xeon(R) Platinum 8352V CPU @ 2.10GHz}.
We test RAG in a multi-turn dialogue setting, where retrieval is required at each interaction.
Initially, the LLM is provided with an instruction comprising 100 tokens. 
The LLM first generates a token to determine the necessity of retrieval. 
After that, a query formulated from the same 100 tokens is sent to the retriever, which retrieves a document containing 30 tokens. 
This document is subsequently concatenated to the initial instruction, and the LLM generates an output of 10 tokens as the response. 
Batch size is set to 1 during the test.
Prior to actual testing, the model undergoes 10 warm-up cycles. 
We assume mandatory retrieval in each iteration and that the instruction and query are of equivalent length. 
However, it is notable that most existing methodologies do not circumvent the step of query rewriting. 
In Figure~\ref{fig:exp:analysis:efficiency}(a), we conduct five dialogue rounds and report the cumulative time expended across these dialogues.
For \ours, an additional token is generated in each round of dialogue for retrieval purposes.

\subsubsection{Impacts on Generation and Retrieval}
\label{app:rg_single:experiments:impact}
\begin{figure}[!ht] 
    \centering
    \scalebox{0.65}{
    \includegraphics[width=1\linewidth]{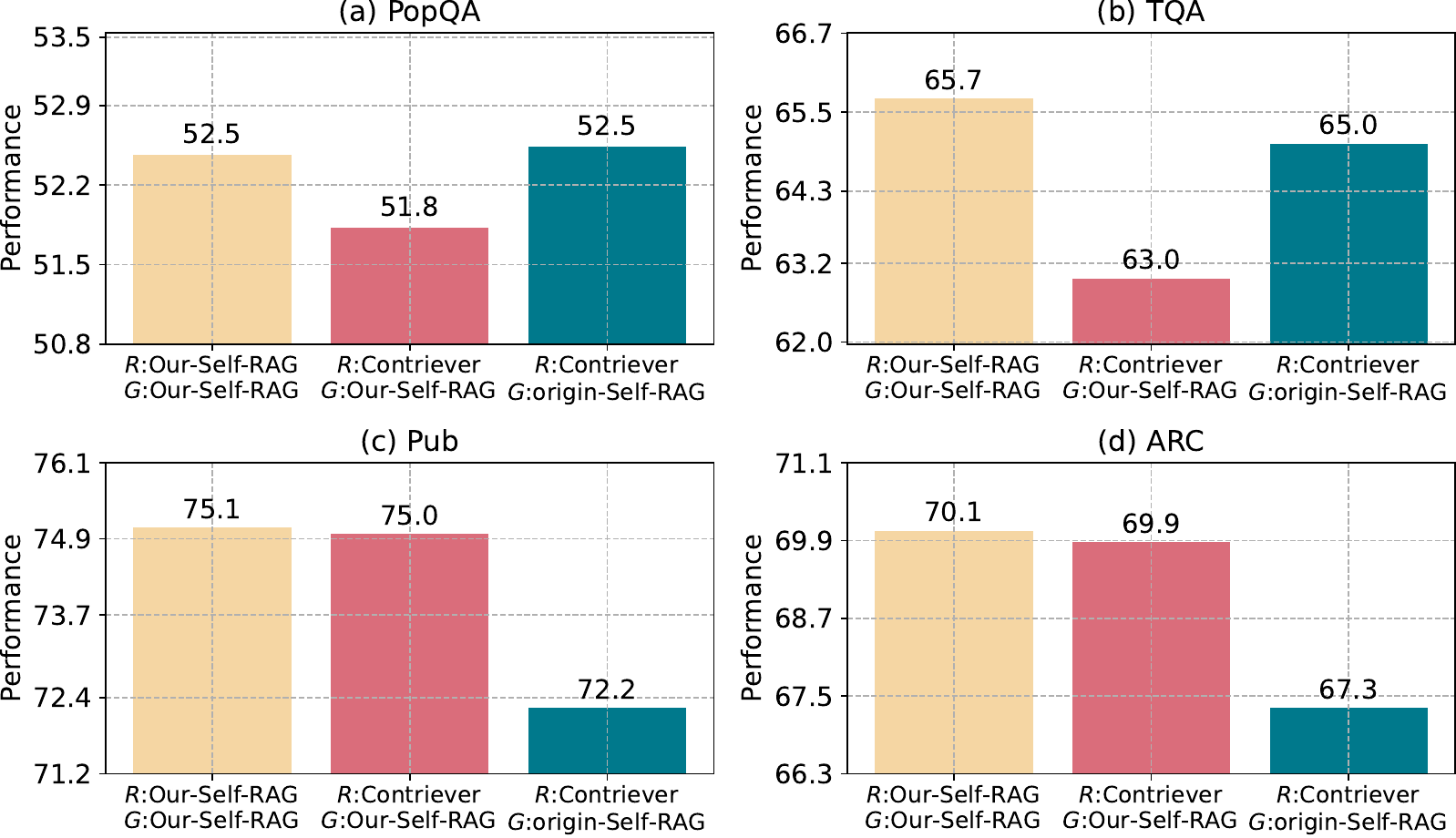} 
    }
    \caption{Performance comparison of three configurations across four datasets in a RAG task.
    Each bar represents a specific setup, with `R' denoting the model used for retrieval and `G' indicating the model used for generation.}
    \label{fig:exp:analysis:impact:rag}
\end{figure} 
$\textbf{R}\rightarrow \textbf{G Task.}$
\textit{Same Retriever but different Generator:}
We employ Contriever as the retriever. 
Specifically, the document ranks are obtained from it.
Comparing the red and blue bars in Figure~\ref{fig:exp:analysis:impact:rag}, we observe a slight decrease in performance of approximately 0.7 points on PopQA and a decrease of 2.0 points on TriviaQA.
Conversely, performance improved by 2.8 and 2.6 points on the Pub and ARC datasets respectively, likely due to joint training, consistent with findings from RA-DIT~\citep{iclr24_RA-DIT} and Grit~\citep{arxiv2024_GritLM}.
\textbf{Overall, \ours~does not adversely affect the generative capabilities of LLMs.}
\textit{Same Generator but different Retriever:}
We assess the performance of different retrievers by examining the output of the generator.
Observing the yellow and red bars in Figure~\ref{fig:exp:analysis:impact:rag}, we note an increase in performance of 0.7 and 2.7 points on the PopQA and TriviaQA datasets, respectively.
Slight gains were also observed in the other datasets, indicating that \textbf{\ours~effectively enhances the retrieval capabilities of LLMs.}

\subsubsection{Ablation}
\label{app:rg_single:expriments:ablation}

\begin{figure*}[] 
    \centering
    \scalebox{0.9}{
    \includegraphics[width=1\linewidth]{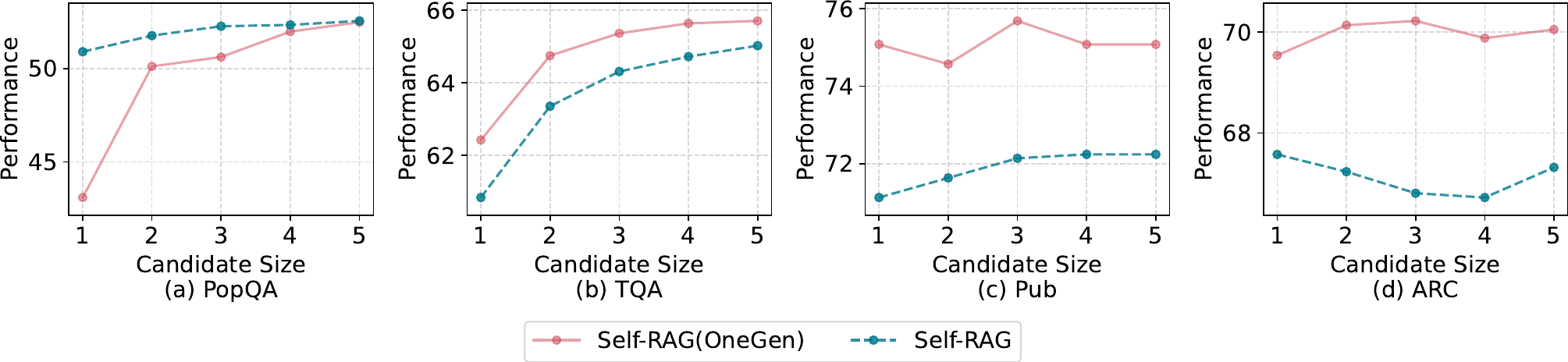} 
    }
    \caption{
    Performance of Self-RAG on four datasets at different candidate sizes from Top 1 to 5.
    }
    \label{fig:exp:analysis:ablation:rag}
    \vspace{-3mm}
\end{figure*} 

\textbf{Top-K.}
Self-RAG is capable of autonomously evaluating its retrieval outcomes.
Specifically, if a retriever produces the Top-K results, Self-RAG processes and assesses each result independently, necessitating K evaluations.
Based on these assessments, it selects the highest-scoring result.
We explore values of K ranging from 1 to 5, with the findings depicted in Figure~\ref{fig:exp:analysis:ablation:rag}. 
The observed trend is consistent between Self-RAG and Self-RAG (\ours).
Performance on the PopQA and TriviaQA datasets increases with larger K, demonstrating the effectiveness of the self-assessment mechanism.


\clearpage
\section{$\textbf{R}\rightarrow \textbf{G}$ Task: RAG for Multi-hop QA}
\label{app:rg_multi}
\subsection{Training Details}
\label{app:rg_multi:training}

\textbf{Data Reconstruction.}
Our data reconstruction process consists of two stages. 
First, we use Qwen2-72B with our designed prompts illustrated in Figure~\ref{prompt:annotation:multi_hop} to further annotate the data from the HotpotQA~\citep{emnlp18_hotpotqa} and 2WIKI~\citep{coling20_2wiki} datasets.
Specifically, each query is decomposed into a sequence of sub-queries, and each sub-query is matched with the corresponding documents.
After annotation, a modified example is as follows:
\begin{tcolorbox}
\textbf{Input:}\\
\texttt{<Instruction>.}
Is Yangzhong or Cenxi located in the east of Guangxi, People's Republic of China?\\

\textbf{Output:}\\
\textcolor{brown}{First, In which province is Yangzhong located?}
\textcolor{violet}{\texttt{[RQ]}}\\
\texttt{<paragraph>}wiki title: Yangzhong\\
content: Yangzhong is a beautiful city. Yangzhong is also a county-level city under the administration of Zhenjiang, Jiangsu province, China.  It is the easternmost county-level division of Zhenjiang City.\texttt{</paragraph>}\\
\textcolor{brown}{Yangzhong is located in Jiangsu province, China.}\\

\textcolor{brown}{Second, Is Cenxi located in the east of Guangxi, People's Republic of China?}
\textcolor{violet}{\texttt{[RQ]}}\\
\texttt{<paragraph>}wiki title: Cenxi\\
content: Cenxi is a county-level city under the administration of Wuzhou City, in the east of Guangxi, People's Republic of China.\texttt{</paragraph>}\\
\textcolor{brown}{Yes, Cenxi is located in the east of Guangxi, People's Republic of China.}\\

\textcolor{brown}{Therefore, Cenxi is the city located in the east of Guangxi, People's Republic of China, not Yangzhong.  \\
$<$FINAL-ANSWER$>$Cenxi$<$/FINAL-ANSWER$>$}
\end{tcolorbox}
In this example, the tokens highlighted in brown are involved in the calculation of the $\mathcal{L}_g$ loss, while those highlighted in purple are involved in the calculation of the $\mathcal{L}_r$ loss. 
Special tokens used include \texttt{<paragraph>}, \texttt{</paragraph>}, and \texttt{[RQ]}.

Since both 2WIKI and HotpotQA provide candidate documents for each query and each document is segmented into sentences, our strategy for document representation is to perform a forward pass on each document. 
If a document contains $n$ sentences, this results in $n$ representations. 
Specifically, we append the \texttt{[RD]} token to each sentence within the document. 
For illustration, consider the above query ``In which province is Yangzhong located?'' to explain how we select positive and negative samples. 
Suppose we have two documents:
\begin{figure}[h!]
\begin{tcolorbox}
\textbf{Doucment 1:}\\
Yangzhong is a beautiful city.
\textcolor{violet}{\textbf{\texttt{[RD]}}}$_\texttt{1}$
Yangzhong is also a county-level city under the administration of Zhenjiang, Jiangsu province, China. 
\textcolor{violet}{\textbf{\texttt{[RD]}}}$_\texttt{2}$
It is the easternmost county-level
division of Zhenjiang City.
\textcolor{violet}{\textbf{\texttt{[RD]}}}$_\texttt{3}$\\ \\
\textbf{Doucment 2:}\\
Cenxi is a county-level city under the administration of Wuzhou City, in the east of Guangxi, People’s Republic of China.\textcolor{violet}{\textbf{\texttt{[RD]}}}$_\texttt{4}$
\end{tcolorbox}
\vspace{-3mm}
\caption{A case for data reconstruction of positive document and negative document about Multi-hop QA. Add \texttt{[RD]} token at the end of each sentence in every document.}
\label{app:fig:rg:multi_hop:doc:case}
\end{figure}

To distinguish them, we append an ID to each \texttt{[RD]} token, though in practice they are identical. 
The positive samples for this query are \texttt{[RD]}$_\texttt{2}$ and \texttt{[RD]}$_\texttt{3}$, while the negative samples are \texttt{[RD]}$_\texttt{1}$ and \texttt{[RD]}$_\texttt{4}$. 
A segment containing the expected answer is considered a positive sample; otherwise, it is a negative sample. 
Thus, \texttt{[RD]}$_\texttt{2}$ is expected to retrieve information from \texttt{[RD]}$_\texttt{1}$, and \texttt{[RD]}$_\texttt{3}$ is expected to capture representations from the preceding two sentences.

\textbf{Implementation Details.}
\begin{table}[]
\small
\centering
\begin{tabular}{llll}
\toprule
\multirow{2}{*}{\textbf{BackBone}} & \multicolumn{3}{c}{\textbf{Hyper Parameters}} \\ \cmidrule{2-4} 
                                   & Pos. per Sent.  & Neg. per Pos.  & Max Length \\ \midrule
Llama2-7B                          & 2               & 6              & 1200       \\
Llama3.1-7B                        & 2               & 2              & 1200       \\
Qwen2-1.5B                         & 2               & 4              & 1100       \\
Qwen2-7B                           & 2               & 2              & 1100       \\ \bottomrule
\end{tabular}
\caption{Training hyper parameters for different backbone in Multi-hop QA setting.}
\label{table:app:single_hop:train:hyperparameter}
\end{table}
Following the previous steps, we randomly sampled 10\% of the HotpotQA and 2WIKI training datasets for annotation. 
After the annotation process, we obtained 9,044 samples from HotpotQA and 16,745 samples from 2Wiki, resulting in a total of 25,789 training samples. 
For evaluation, we utilize the complete validation sets of HotpotQA and 2WIKI, as the original test sets do not provide ground truth labels. 
The HotpotQA validation set consists of 7,405 samples, while the 2WIKI validation set comprises 12,576 samples.
We perform comprehensive training on eight A800 machines utilizing the DeepSpeed ZeRO-3 strategy for memory efficiency. 
We set the gradient accumulation to 4, and the batch size per GPU is 2, resulting in a final global batch size of $8\times 2\times 4=64$. 
Training is conducted over 3 epochs with a learning rate set at 2e-5 and a 3\% warm-up period. 
Both $\lambda_g$ and $\lambda_r$ are set at 1.
Other hyper-parameters are shown in Table~\ref{table:app:single_hop:train:hyperparameter}.

\subsection{Evaluation Details}
\label{app:rg_multi:evaluation}
\textbf{Baselines.}
We use a pipeline method as the baseline for the Multi-hop QA setting, where the pipeline alternates between the retriever and the generator.
For the retriever, we employ the untrained Contriever~\citep{tmlr22_contriever}, which is consistent with the retrievers used in RQ-RAG~\citep{arxiv24_RQ-RAG} and Self-RAG~\citep{iclr24_selfrag}.
For the generator, we train on data constructed as described in Appendix~\ref{app:rg_multi:training}.
The only difference from \ours~is the omission of the $\mathcal{L}_r$ loss, meaning that the \texttt{[RQ]} token used as input to the LLM is not involved in optimization.
During the inference, the generation of the \texttt{[RQ]} token by the LLM indicates the need to call the retriever.
We input each generated sub-query into the retriever for retrieval.

\textbf{Evaluation Metrics.}
We utilize the code provided by the original papers~\citep{emnlp18_hotpotqa, coling20_2wiki} to evaluate the generation of both the pipeline and our method using F1 and EM metrics.
For retrieval evaluation, since the LLM used by the pipeline and the LLM trained with \ours~have different model parameters, the queries generated for the same question differ.
Thus, we report the retrieval results for queries generated by Contriever from the pipeline, and the retrieval results for queries generated using \ours.
Specifically, for a given query requiring two steps of reasoning, two sub-queries are generated, each retrieving a relevant document. 
If these retrieved documents match the ground truth, the retrieval is considered correct; otherwise, it is considered incorrect.

\subsection{Experiments}
\label{app:rg_multi:experiments}

\subsubsection{Ablation}
\label{app:rg_multi:experiments:ablation}
\begin{figure*}[] 
    \centering
    \scalebox{1.0}{
    \includegraphics[width=1\linewidth]{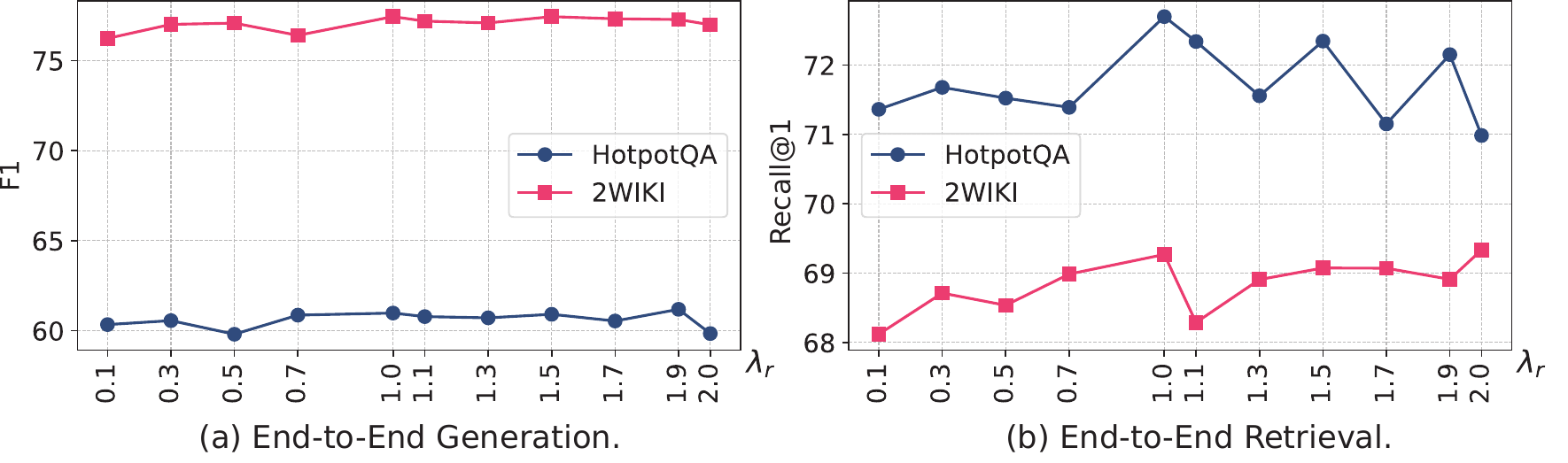} 
    }
    \caption{
    Ablation study for $\lambda_r$ using Qwen2-1.5B.
    }
    \label{app:fig:rg:multi_hop:ablation:lambda}
\end{figure*} 
\textbf{Hyper parameter $\lambda_r$.}
We examine the impact of $\lambda_r$ using the Qwen2-1.5B for Multi-hop QA task.
We search the $\lambda_r$ from $\{0.1,0.3,0.7,1.0,1.3,1.5,1.7,1.9,2.0\}$ and the result are presented in Figure~\ref{app:fig:rg:multi_hop:ablation:lambda}.
We find that \ours~is insensitive to hyper- parameters $\lambda_r$, demonstrating the robustness of the \ours.

\begin{table}[]
\centering
\begin{tabular}{lcccc}
\toprule
\multirow{2}{*}{\textbf{Method}}    & \multicolumn{2}{c}{\textbf{HotpotQA}} & \multicolumn{2}{c}{\textbf{2WIKI}} \\
                                    & EM                & F1                & EM               & F1              \\ \midrule
Llama2-7B + Explicit Query + \ours & \textbf{54.82}    & \textbf{67.93}    & \textbf{75.02}   & \textbf{78.86}  \\
Llama2-7B + Implicit Query + \ours & 51.02             & 63.27             & \textbf{72.92}   & \textbf{79.23}  \\ \midrule
Llama2-7B + Contriever              & 52.83             & 65.64             & 70.02            & 74.35           \\ \bottomrule
\end{tabular}
\caption{Ablation study for using the implicit query.}
\label{table:exp:ablation:implicit_query}
\end{table}
\textbf{Implicit Query.}
Here, we remove the explicit query to observe the performance.
Take the previous query ``\texttt{In which province is Yangzhong located?}'' in the of \textit{Data Reconstruction} Appendix~\ref{app:rg_multi:training} as an example, we replace the explicit query with the implicit query ``\texttt{The question is:}''.
Table~\ref{table:exp:ablation:implicit_query} shows the result.
We find that \ours~can get a good performance under the implicit query settings.

\begin{figure}[]
\begin{tcolorbox}
You excel in question decomposition. Starting with a question, related documents, and the final answer corresponding to the question, your task is to break down the question into sub-questions. Each sub-question should be connected to relevant documents, leading to the generation of answers for each sub-question. It's important to consider potential connections between these sub-questions. \\
Here’s an example input:\\
$<$QUESTION$>$ \\
Are both magazines, the Woman's Viewpoint and Pick Me Up, British publications?\\
$<$/QUESTION$>$\\
$<$ANSWER$>$
no
$<$/ANSWER$>$\\
$<$RELATED-DOC-1$>$\\
wiki title: Pick Me Up (magazine)\\
content: Pick Me Up! is a British weekly women's magazine that is published through the IPC Media group.\\
$<$/RELATED-DOC-1$>$\\
$<$RELATED-DOC-2$>$\\
wiki title: Woman's Viewpoint (magazine)
content: The Woman's Viewpoint was a woman's magazine founded in Texas in 1923 and published by Florence M. Sterling. The magazine was progressive and ran from 1923 to 1927.\\
$<$/RELATED-DOC-2$>$\\
\\
Here's an example output for the given input:\\
$<$SUB-QUESTION-1$>$\\
Which country is the magazine Woman's Viewpoint published in?\\
$<$/SUB-QUESTION-1$>$\\
$<$CORRESPONDING-DOC-1$>$\\
RELATED-DOC-2\\
$<$/CORRESPONDING-DOC-1$>$\\
$<$SUB-ANSWER-1$>$\\
The magazine 'Woman's Viewpoint' was published in Texas, a state located in the United States.\\
$<$/SUB-ANSWER-1$>$\\
$<$SUB-QUESTION-2$>$\\
Which country is the magazine Pick Me Up published in?\\
$<$/SUB-QUESTION-2$>$\\
$<$CORRESPONDING-DOC-2$>$\\
RELATED-DOC-1\\
$<$/CORRESPONDING-DOC-2$>$\\
$<$SUB-ANSWER-2$>$\\
The magazine Pick Me Up! is published in the United Kingdom.\\
$<$/SUB-ANSWER-2$>$\\
$<$FINAL-ANSWER$>$\\
No, they are not both British publications. ``Pick Me Up!'' is indeed a British magazine published in the United Kingdom. However, ``Woman's Viewpoint'' was published in Texas, which is in the United States, so it is an American publication.\\
$<$/FINAL-ANSWER$>$\\
\\
Here is an input you need to process:\\
$<$QUESTION$>$
\texttt{\{question\}}
$<$/QUESTION$>$\\
$<$ANSWER$>$
\texttt{\{answer\}}
$<$/ANSWER$>$\\
\texttt{\{related\_doc\}}\\
\\
Please format your output as shown above and refrain from including any additional content.
\end{tcolorbox}

\caption{Prompt for constructing data in Multi-hop QA setting, using Qwen72B.}
\label{prompt:annotation:multi_hop}
\end{figure}

\clearpage
\section{$\textbf{G}\rightarrow \textbf{R}$ Task: Entity Linking}
\label{app:gr}
This section presents the details of the $\textbf{G}\rightarrow \textbf{R}$ task, including the detailed construction of the data, training details, inference details, and additional experimental results.
\subsection{Training Details}
\label{app:gr:training}
A comprehensive example of training Entity Linking is illustrated in Figure~\ref{fig:app:training} (a).

\textbf{Data Reconstruction.}
Here, we outline the adaptations we implemented in the training dataset for Entity Linking. 
Initially, we demonstrate how LLMs can be utilized to perform the Mention Detection task in a generative fashion. 
As shown in the following diagram, the input comprises specific extraction instructions alongside the sentence targeted for extraction, and the output is a sentence annotated accordingly. 
All outputs contribute to the calculation of the generative loss.
\begin{tcolorbox}
\textbf{Input:}
\texttt{<Instruction>.}
Steve Jobs founded Apple Inc.

\textbf{Output:}
\textcolor{brown}{
\texttt{<MENTION>}
Steve Jobs\texttt{</MENTION>} founded \texttt{<MENTION>}Apple Inc\texttt{</MENTION>}.}
\end{tcolorbox}

For this case, we append two special tokens, \texttt{[RQ]} and \texttt{[CON]}, subsequent to each \texttt{</MENTION>} token. 
The \texttt{[RQ]} token is intended to prompt the model to extract semantic information from the preceding mention, whereas the \texttt{[CON]} token aids in the LLMs' generation of subsequent content. 
It should be noted that only the \texttt{[RQ]} token is considered in the computation of representative loss, while the \texttt{[CON]} token remains involved in the generative loss calculation.
\begin{tcolorbox}
\textbf{Input:}
\texttt{<Instruction>.}
Steve Jobs founded Apple Inc.

\textbf{Output:}
\textcolor{brown}{\texttt{<MENTION>}Steve Jobs\texttt{</MENTION>}}
\textcolor{violet}{\texttt{[RQ]}}
\textcolor{violet}{\texttt{[CON]}}
\textcolor{brown}{founded} 
\textcolor{brown}{\texttt{<MENTION>}Apple Inc\texttt{</MENTION>}}
\textcolor{violet}{\texttt{[RQ]}}
\textcolor{violet}{\texttt{[CON]}}
\textcolor{brown}{.}
\end{tcolorbox}
The computation of representative loss necessitates the use of both positive and negative examples. 
Here is a set of positive and negative examples about \texttt{Steve Jobs}:

\begin{figure}[h]
\begin{tcolorbox}
\textbf{Positive:}

Steven Paul Jobs (February 24, 1955 – October 5, 2011) was an American businessman, inventor, and investor best known for co-founding the technology giant Apple Inc.
\textcolor{violet}{\texttt{[RD]}}

\textbf{Negative:}

Steve Jobs is a 2015 biographical drama film directed by Danny Boyle and written by Aaron Sorkin. A British-American co-production, it was adapted from the 2011 biography by Walter Isaacson and interviews conducted by Sorkin.
\textcolor{violet}{\texttt{[RD]}}
\end{tcolorbox}
\vspace{-3mm}
\caption{A case for data reconstruction of positive document and negative document about Entity Linking. Add \texttt{[RD]} token at the end of each document.}
\label{app:fig:gr:el:doc:case}
\end{figure}
\vspace{-2mm}
In these examples, the \texttt{[RQ]} token directs the LLMs to extract semantic information from the current document. 
Only the \texttt{[RQ]} token participates in the representative loss computation, with the other elements excluded from this process.

\textbf{Implementation Details.}
We conducted our training using the AIDA dataset combined with 1\% of Wikipedia data, yielding a total of approximately 68k samples. 
The training is performed on eight A800 machines, leveraging the DeepSpeed ZeRO-3 strategy to optimize memory utilization. 
We configured the maximum sequence length at 1300 and established a batch size of 5 per GPU, which resulted in an overall global batch size of 40. 
Each sentence in the dataset contained multiple \texttt{[RQ]} tokens, from which we randomly selected one token per sentence for optimization, paired with one positive and four negative samples. 
The negative samples were shared across the batch. 
The training regimen included 5k steps with a learning rate of 5e-6.

\subsection{Inference Details}
\label{app:gr:inference}
Algorithm~\ref{algo:el} presents the pseudocode for Entity Linking inference.
Figure~\ref{fig:app:inference}(a) provides a schematic representation of the Entity Linking inference process.

\textbf{Candidate Construction.}
Our consolidated entity library comprises five distinct repositories:
\begin{itemize}
\item{
Utilizing ReFinED~\citep{naacl22_refined}, we generate \textit{Candidate Repository 1} from the training datasets based on annotated mentions.
}
\item{
We utilize \textit{Candidate Repository 2} for the test datasets, sourced from ChatEL~\citep{coling24_ChatEL}, which is annotated by REL~\citep{sigir20_REL} and BLINK~\citep{emnlp20_BLINK}.
}
\item{
\textit{Candidate Repository 3} is included, 
provided by~\cite{emnlp11_aida}.
}
\item{
Through the Tool from~\cite{arxiv22_candidatetool}, we extract the top 10 challenging candidates from both the AIDA training and test datasets within the Wikidata repository, resulting in \textit{Candidate Repository 4}.
}
\item{
\textit{Candidate Repository 5} is created by applying ReFinED to mentions extracted by our method.
}
\end{itemize}

\subsection{Evaluation Details}
\label{app:gr:evaluation}
\textbf{Baselines.}
For EL, we employ REL 2019~\citep{sigir20_REL}, Neural EL~\citep{conll18_neural-el}, GENRE~\citep{iclr21_GENRE} and ReFinED~\citep{naacl22_refined} as our baselines.
For MD, we expand EL baselines to include LLMs such as Qwen-7B-chat, Qwen-14B-chat~\citep{arxiv23_qwen}, Llama2-7B-chat, and Llama3-7B-chat.
For ED, we expand EL baselines to include ChatEL~\citep{coling24_ChatEL} and EntGPT~\citep{arxiv24_EntGPT}.
ChatEL transforms the ED task into a question-answering task, utilizing prompts to facilitate GPT-4's execution of the task.
Conversely, EntGPT involves fine-tuning GPT-3.5~\citep{ChatGPT-OpenAI} using specially constructed datasets.

\textbf{Evaluation Setup, Datasets, and Metric.}
For EL task, we utilize the ELEVANT~\citep{emnlp22_ELEVANT} tool for evaluation across seven datasets: the in-domain AIDA-CoNLL (AIDA) and six out-of-domain datasets, MSNBC~\citep{emnlp07_msnbc}(MSN), KORE50~\citep{cikm12_k50}(K50), N3-Reuters-128~\citep{lerc14_Reu}(REU), SpotLight (SPOT), OKE Challenge 2015~\citep{eswc15_oke15}(O15), and OKE Challenge 2016~\citep{eswc16_oke16} (O16), using the Micro F1 metric to evaluate in-KB entities.
The same datasets and metrics are applied to the MD task.
For ED task, following the ChatEL~\citep{coling24_ChatEL}, we extend our evaluation to include the RSS~\citep{lerc14_Reu} and ACE04 datasets, maintaining the use of the Micro F1 metric.
In assessing candidate entities for each mention in EL task, our methodology leverages a proprietary entity repository of 1.25M entities derived from Wikipedia and Wikidata, which includes numerous indistinguishable entities. 
For baseline comparisons, the ELEVANT~\citep{emnlp22_ELEVANT} tool is also employed.

\renewcommand{\thealgorithm}{2}
\label{algo:el}
\begin{algorithm*}
\caption{Entity Linking Inference}
\begin{algorithmic}[1]
\Require 
\Statex LLM trained with OneGen, denoted as $\hat{f}(\cdot)$
\Statex LLM without the LM-Head, denoted as $f(\cdot)$
\Statex Pre-cached document vector library $Emb_{doc}$
\Statex Instruction with text to be extracted $x$
\Statex Cosine similarity computation function \texttt{CosineSimilarity()}
\Statex Function to sort and return indices \texttt{Top1()}

\Ensure
\Statex Text with marked mentions $History$
\Statex List of entities id corresponding to the mentions in the text $EntityList$

\Statex\hspace*{-\algorithmicindent}\hrulefill
\State $History \gets x$
\State $EntityList \gets []$
\State $NextToken \gets \hat{f}(History)$
\While{$NextToken \notin \texttt{Terminator}$}
    \State $History \gets History \cup \{NextToken\}$
    \If{$role(NextToken) = \texttt{RET}$}\textcolor{blue}{{\Comment{Retrieval on demand}}}
        \State $scores \gets \texttt{CosineSimilarity}(f(History), Emb_{doc})$
        \State $EntityList\gets EntityList \cup \{\texttt{Top1}(scores)\}$
        \State $History \gets History \cup \{\texttt{[CON]}\}$
    \EndIf
    \State $NextToken \gets \hat{f}(History)$ \textcolor{blue}{\Comment{Generation}}
\EndWhile
\State \Return $History, EntityList$
\end{algorithmic}
\end{algorithm*}
\subsection{Experiments}
\label{app:gr:experiments}
\subsubsection{Efficiency Settings}
\label{app:gr:experiments:efficiency}
We conduct tests on a single 40GB A100 card, with an \texttt{Intel(R) Xeon(R) Platinum 8352V CPU @ 2.10GHz}.
For the configuration of the pipeline, it involves two LLMs: one for extracting mentions from sentences and another for linking based on the extracted mentions. 
The extraction process is driven by an instruction that includes the target sentence and a specific extraction instruction, resulting in a sentence annotated with annotation.
The linking process is structured as a question-answering (QA) task, utilizing a prompt composed of the sentence, a linking instruction, a list of four candidates.  
The output for linking is set to a single token. 
In the OneGen configuration, ours allow for direct retrieval during the generation process. 
Consequently, the number of output tokens surpasses those from the pipeline's extraction output by an increment of $2n$ tokens, where $n$ represents the number of mentions in the sentence.
Specifically, the sentence intended for extraction is limited to 1000 tokens, while the instructions for both extraction and linking are capped at 15 tokens each, and each candidate description is confined to 30 tokens.
\subsubsection{Ablation}
\label{app:gr:experiments:ablation}

\begin{table}[]
\centering
\small
\begin{tabular}{lcr}
\toprule
\textbf{Method}           & \textbf{Recall Strategy} & \textbf{AVG.}         \\ \midrule
ReFinED          & ReFinED       & 60.8        \\
Llama2$_{\texttt{7B}}$+\ours & ReFinED       & {\underline {61.8}}  \\
Llama2$_{\texttt{7B}}$+\ours & Ours          & \textbf{64.0} \\ \bottomrule
\end{tabular}
\caption{Ablation study results of recall strategies for entity linking, reporting average F1 scores across seven datasets.}
\label{table:exp:ablation:gr:recall}
\end{table}
\textbf{Recall Strategy for candidate sets.}
In the EL task, once a mention is extracted, it is necessary to select the corret entity corresponding to the mention from a candidate entity repository.
Our previous method build a challenging repository that allow \ours~to choose from 1.25M entities without additional recall strategy, achieving 100\% recall on the test set.
For ablation, we use the ReFinED recall strategy, which maps mentions to 30 potential entities using a rule-based dictionary, with a recall rate of 96\% on the test set. 
Our evaluations across seven EL datasets (reported in Table~\ref{table:exp:ablation:gr:recall}) demonstrate that \ours~consistently outperforms ReFinED, indicating that improvements in the retrieval process significantly enhance EL performance.


\begin{table*}[!t]
\centering
\small
\scalebox{0.73}{
\begin{tabular}{lcccccccccccc}
\toprule
\multicolumn{1}{l}{\textbf{Method}} & \multicolumn{1}{c}{\textbf{Training Data}} & \textbf{Cand. Size} & K50 & OKE15 & OKE16 & REU & RSS & ACE04 & MSN & WIKI & AQU & \textbf{AVG.} \\ \midrule
\multicolumn{13}{c}{\textit{Baselines}} \\ \midrule
REL                       & - &  \multirow{4}{*}{\begin{tabular}[c]{@{}c@{}}$<$30 \end{tabular}} & 61.8          & 70.5          & 74.9          & 66.2          & 68.0          & 89.7          & 93.0          & 78.3          & 88.1          & 76.7          \\
Neural EL & - &  & 76.7 & 78.3          & 67.7          & 72.0          & 88.0          & 92.0          & 74.0          & 88.0          & 77.1          \\
GENRE                     & WIKI 6M+AIDA                 &   & 54.2          & 64.0          & 70.8          & 69.7          & 70.8          & 84.8          & 78.0          & 82.3          & 84.9          & 73.3          \\
ReFinED                   & WIKI 6M+AIDA                 &                                                                                     & 56.7          & 78.1          & 79.4          & 68.0          & 70.8          & 86.4          & 89.1          & {\underline{84.1}}    & {\underline{86.1}}    & 77.6          \\ \midrule
\multicolumn{13}{c}{\textit{LLMs Baselines}}                                                                                                                                                                                                                                                                              \\ \midrule
ChatEL (GPT-4)            & \multicolumn{1}{c}{Prompt}   & \multirow{3}{*}{\begin{tabular}[c]{@{}c@{}}$<$30 \end{tabular}} & \textbf{78.7} & 75.8          & 75.2          & 78.9          & 82.2          & 89.3          & 88.1          & 79.1          & 76.7          & 80.4          \\
EntGPT-P (GPT-3.5)        & \multicolumn{1}{c}{AIDA}     &                                                                                     & 71.6          & 76.7          & 77.0          & 78.5          & 80.8          & 91.8          & 86.7          & 80.8          & 79.1          & 80.3          \\
EntGPT-I (GPT-3.5)        & \multicolumn{1}{c}{AIDA}     &                                                                                     & 75.3          & {\underline{82.5}}    & {\underline{81.9}}    & {\underline{80.8}}    & {\underline{82.5}}    & \textbf{93.7} & {\underline{92.2}}    & 79.1          & \textbf{90.6} & {\underline{84.3}}    \\ \midrule
\multicolumn{13}{c}{\textit{Our Method}}                                                                                                                                                                                                                                                                                \\ \midrule
Llama2$_{\texttt{7B}}$(+\ours)                 & WIKI 60K+AIDA                & 1.25M                                                                               & {\underline{77.0}}    & \textbf{87.5} & \textbf{87.5} & \textbf{85.2} & \textbf{85.3} & {\underline{92.2}}    & \textbf{92.5} & \textbf{85.5} & 86.0          & \textbf{86.5} \\ \bottomrule
\end{tabular}
}
\caption{
Entity Disambiguation InKB micro F1 scores on in-domain and out-of-domain test sets. 
The best value in bold and second best is underlined.
The results of the baselines come from ChatEL~\citep{coling24_ChatEL} and EntGPT~\citep{arxiv24_EntGPT}. 
The dataset used here differs slightly from the dataset used for Entity Linking.
For details, refer to ChatEL.
}
\label{table:app:exp:gr:ed}
\end{table*}
\begin{table*}[!t]
\centering
\small
\scalebox{0.81}{
\begin{tabular}{lcccccccccc}
\toprule
\multicolumn{1}{l}{\multirow{3}{*}{\textbf{Method}}} & \multicolumn{1}{l}{\multirow{3}{*}{\textbf{Training Data}}} & \textbf{In-Domain}     &  & \multicolumn{6}{c}{\textbf{Out-of-domain}}                                                             & \multirow{3}{*}{\textbf{AVG.}} \\ \cmidrule{3-3} \cmidrule{5-10}
\multicolumn{1}{c}{}                       &            & AIDA          &  & OKE15         & OKE16         & REU           & MSN           & SPOT          & KORE50        &                      \\ \midrule
\multicolumn{11}{c}{\textit{Baselines}}                                                                                                                                                                                          \\ \midrule
REL2014                                    &  - & 90.3          &  & 67.2          & 58.8          & 61.8          & 82.6          & 27.7          & {\underline{95.2}}    & 69.1                 \\
REL2019                                    & - & 90.5          &  & 67.5          & 58.8          & 62.3          & {\underline{82.8}}    & 27.7          & {\underline{95.2}}    & 69.3                 \\
Neural EL                                  &  AIDA  & {\underline{95.8}}    &  & {\underline{68.6}}    & 60.3          & 71.4          & 79.3          & 23.4          & 82.1          & 68.7                 \\
GENRE                                 & Wiki 6M + AIDA                   & 85.5          &  & 54.4          & 47.5          & 45.6          & 68.5          & 26.9          & 83.3          & 58.8                 \\
ReFinED                                    & Wiki 6M + AIDA                   & \textbf{95.9} &  & \textbf{70.5} & 61.9          & \textbf{76.7} & \textbf{84.2} & 24.0          & \textbf{95.9} & \textbf{72.7}        \\
\midrule
\multicolumn{11}{c}{\textit{LLMs Baselines using SFT}}                                                                                                                                                                                     \\ \midrule
Qwen$_{\texttt{7B}}$                                  &  \multirow{4}{*}{Wiki 60K + AIDA}     & 85.1          &  & 65.7          & 65.1          & 69.4          & 79.5          & \underline{41.1}          & 94.0          & 71.4                 \\
Qwen$_{\texttt{14B}}$                                   &                                  & 91.2          &  & 66.2          & {\underline{65.8}}    & 71.1          & 78.3          & {\textbf{41.8}}    & 93.4          & {\underline{72.6}}           \\ 
Llama3$_{\texttt{7B}}$                                 &                                  & 88.5          &  & 67.3          & \textbf{65.9} & 69.4          & 75.6          & 37.1          & 92.9          & 71.0                 \\
Llama2$_{\texttt{7B}}$                                  &  & 85.6          &  & {\underline{68.6}}    & 63.9          & {\underline{74.9}}    & 80.6          & 31.4          & 92.9          & 71.1                 \\ \midrule
\multicolumn{11}{c}{\textit{Ours}}                                                                                                                                                                                               \\ \midrule
Llama2$_{\texttt{7B}}$(+\ours)                                  & Wiki 60K + AIDA                  & 88.6          &  & 66.6          & 64.5          & 74.7          & 80.5          & 33.5          & 92.4          & 71.5                 \\ \bottomrule
\end{tabular}
}
\caption{
Mention Detection micro F1 scores on in-domain and out-of-domain test sets. 
The best value in bold and second best is underlined.
}
\label{table:app:exp:gr:md}
\end{table*}



\end{document}